\documentclass{article} 
\usepackage{iclr2027_conference,times}

\newif\ifarxiv
\newif\ifcameraready
\arxivtrue
\camerareadyfalse

\usepackage{amsmath,amsfonts,bm}

\def\eqref#1{equation~\ref{#1}}

\def\1{\bm{1}}

\DeclareMathAlphabet{\mathsfit}{\encodingdefault}{\sfdefault}{m}{sl}
\SetMathAlphabet{\mathsfit}{bold}{\encodingdefault}{\sfdefault}{bx}{n}

\definecolor{customblue}{RGB}{53, 111, 176}
\definecolor{customgreen}{RGB}{53, 134, 74}
\definecolor{customamber}{RGB}{235, 153, 63}
\definecolor{customred}{RGB}{215, 97, 96}
\definecolor{customslate}{RGB}{122, 127, 135}
\definecolor{customteal}{RGB}{44, 176, 192}
\definecolor{custommagenta}{RGB}{201, 125, 182}

\definecolor{customviolet}{RGB}{120, 128, 205}
\definecolor{customink}{RGB}{49, 53, 61}

\definecolor{customblueLt2}{RGB}{153, 172, 207}
\definecolor{customblueLt1}{RGB}{107, 141, 193}
\definecolor{customblueDk1}{RGB}{0, 82, 142}
\definecolor{customblueDk2}{RGB}{0, 57, 101}
\definecolor{customgreenLt2}{RGB}{153, 191, 158}
\definecolor{customgreenLt1}{RGB}{103, 163, 114}
\definecolor{customgreenDk1}{RGB}{0, 104, 44}
\definecolor{customgreenDk2}{RGB}{0, 75, 29}
\definecolor{customamberLt2}{RGB}{255, 234, 217}
\definecolor{customamberLt1}{RGB}{255, 191, 131}
\definecolor{customamberDk1}{RGB}{201, 120, 15}
\definecolor{customamberDk2}{RGB}{164, 95, 0}
\definecolor{customredLt2}{RGB}{248, 182, 177}
\definecolor{customredLt1}{RGB}{236, 139, 134}
\definecolor{customredDk1}{RGB}{184, 60, 65}
\definecolor{customredDk2}{RGB}{149, 31, 42}
\definecolor{customslateLt2}{RGB}{187, 190, 196}
\definecolor{customslateLt1}{RGB}{154, 158, 166}
\definecolor{customslateDk1}{RGB}{92, 97, 106}
\definecolor{customslateDk2}{RGB}{66, 71, 79}
\definecolor{customtealLt2}{RGB}{185, 235, 243}
\definecolor{customtealLt1}{RGB}{121, 206, 218}
\definecolor{customtealDk1}{RGB}{0, 143, 157}
\definecolor{customtealDk2}{RGB}{0, 113, 124}
\definecolor{custommagentaLt2}{RGB}{244, 203, 232}
\definecolor{custommagentaLt1}{RGB}{224, 163, 208}
\definecolor{custommagentaDk1}{RGB}{171, 91, 152}
\definecolor{custommagentaDk2}{RGB}{139, 65, 123}
\definecolor{customvioletLt2}{RGB}{195, 195, 236}
\definecolor{customvioletLt1}{RGB}{158, 161, 223}
\definecolor{customvioletDk1}{RGB}{85, 98, 178}
\definecolor{customvioletDk2}{RGB}{55, 72, 145}
\definecolor{custominkLt2}{RGB}{107, 109, 114}
\definecolor{custominkLt1}{RGB}{77, 80, 87}
\definecolor{custominkDk1}{RGB}{44, 48, 58}
\definecolor{custominkDk2}{RGB}{44, 48, 57}

\usepackage[colorlinks=true,allcolors=customblue]{hyperref} 
\usepackage{url}
\usepackage{graphicx}
\usepackage{booktabs}
\usepackage{tabularx}
\usepackage{array}
\usepackage{wasysym}
\usepackage{amssymb} 
\usepackage{pifont}
\usepackage{wrapfig}
\usepackage{booktabs}
\usepackage{xcolor}  
\usepackage{caption}
\usepackage{tabularx}
\usepackage{array}
\usepackage[table]{xcolor}
\usepackage[skins]{tcolorbox}

\newcolumntype{Y}{>{\raggedright\arraybackslash}X}

\title{\method{}work: Offline Rewriting of Robot Data with Verified Local Stitches}

\author{
Juntao Ren\thanks{Correspondence to: \texttt{juntao.ren@stanford.edu}} \\
Stanford University \\
\And
Yifan Hou \\
NVIDIA \\
\And
Shuran Song \\
Stanford University
}

\newcommand{\takeaway}[1]{%
  \begin{tcolorbox}[
    enhanced,
    colback=customblue!12,
    coltext=black,
    colframe=customblue,
    arc=3pt,
    boxrule=0.6pt,
    boxsep=0pt,
    left=7pt,right=7pt,top=5pt,bottom=5pt,
    before skip=6pt,after skip=6pt
  ]
    \textcolor{customblueDk1}{\textbf{Takeaway.}} #1
  \end{tcolorbox}%
}

\newcommand{\tabsem}[1]{%
  \textcolor{gray}{\scriptsize $\pm #1$}%
}

\newcommand{\method}{\textsc{Needle}}

\definecolor{lenshadegreen}{RGB}{53, 134, 74}   
\definecolor{lenshadered}{RGB}{215, 97, 96}     

\ifarxiv
  \ifcameraready
    \PackageError{paper}{Choose either arXiv or ICLR camera-ready}{Set one version switch to false.}
  \fi
  \iclrfinalcopy 
\else
  \ifcameraready
    \iclrfinalcopy 
  \fi
\fi
\hypersetup{
  pdftitle={NEEDLEWORK: Offline Rewriting of Robot Data with Verified Local Stitches}
}
\ificlrfinal
  \hypersetup{pdfauthor={Juntao Ren, Yifan Hou, Shuran Song}}
\else
  \hypersetup{pdfauthor={}}
\fi

\begin{document}

\maketitle
\ifarxiv
  \lhead{} 
\fi

\begin{abstract}
Robot demonstrations may contain useful behavior even when individual episodes are inefficient or unsuccessful. Trajectory stitching offers a way to compose these behaviors into improved training data, but identifying useful connections and verifying their feasibility is difficult in high-dimensional robot data, where many prior methods rely on low-dimensional state representations. We introduce \method{}, an offline dataset-augmentation algorithm that addresses these challenges by adding short, verified action bridges between recorded observations in high-dimensional robot demonstrations. First, \method{} identifies and creates connections that bypass suboptimal detours, broaden action coverage, and augment the original dataset with failed trajectories, using only RGB images, proprioception, and episode-level outcomes, without new environment interaction or privileged object state. Next, we present a sampling technique that incorporates accepted bridges into policy training without synthesizing intermediate images or discarding the original demonstrations, allowing policies to learn alternative actions while retaining the original dataset's coverage. On real-robot tasks, \method{} improves success rate over the strongest baseline on each task by an average of 21 percentage points. Videos and supplementary materials are on \href{https://needle-work.github.io/}{needle-work.github.io}.
\end{abstract}

\section{Introduction}
\label{sec:introduction}

\begin{wrapfigure}{r}{0.52\textwidth}
    \vspace{-1.4\baselineskip}
    \centering
    \includegraphics[
        width=\linewidth
    ]{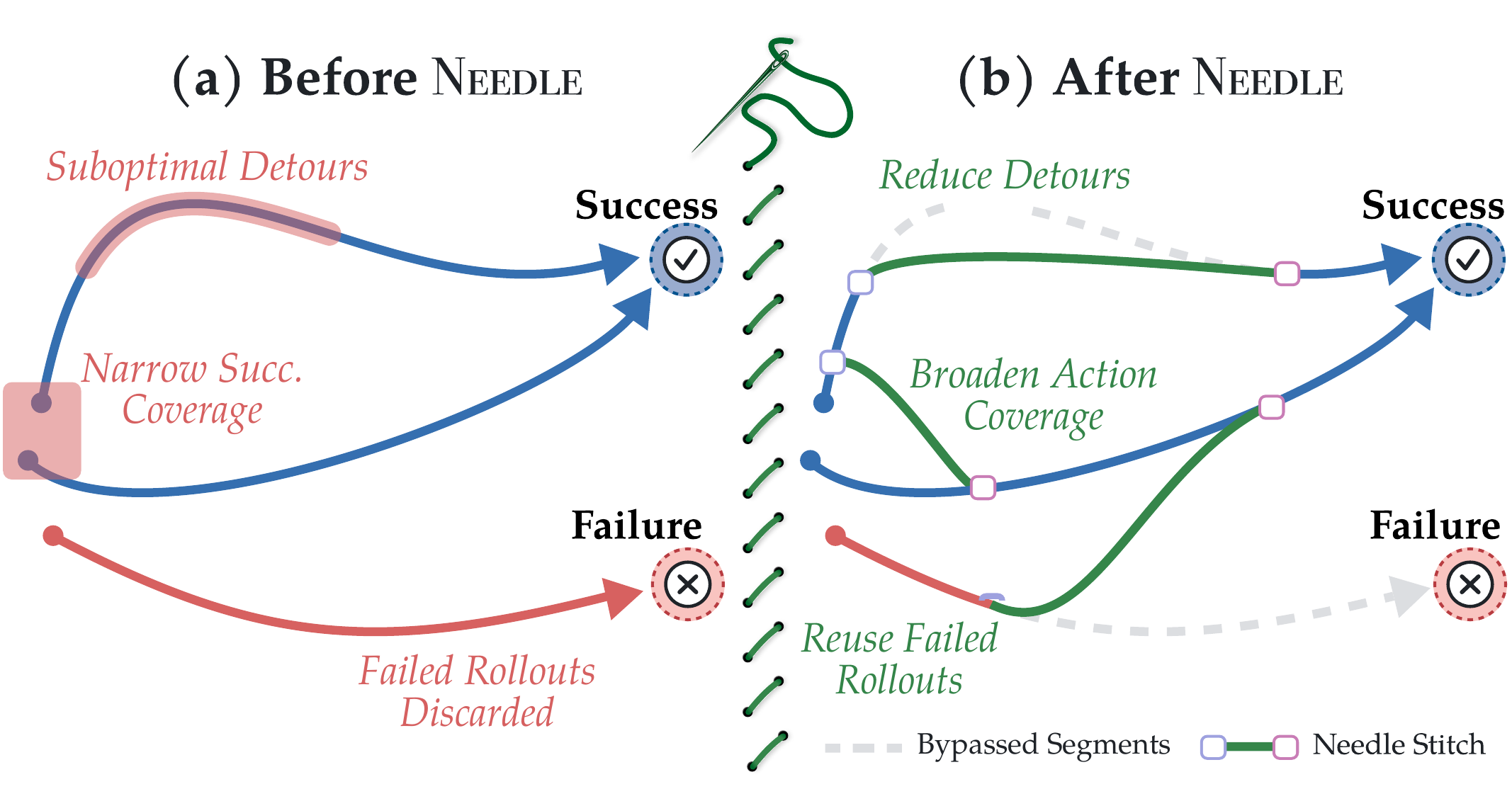}
    \vspace{-2\baselineskip}
    \caption{\method{} produces synthetic bridges within trajectories that bypass inefficient detours, cross-trajectory stitches that discover more successful ways to complete the task, and recover useful behavior from failed episodes.
    }
    \label{fig:teaser}
    \vspace{-1.4\baselineskip}
\end{wrapfigure}

The best robot behavior may not exist in any single trajectory of a dataset. Consider a bimanual laundry-folding task. One demonstration may fold the sleeves reliably but require several corrective grasps to complete the bottom fold; another may fold the sleeves poorly but fold the bottom cleanly. Neither demonstration alone completes the task well, but synthesizing a connection between the two good segments would produce a demonstration better than either individually.

This example illustrates one of three opportunities offered by \textbf{trajectory stitching} (Fig.~\ref{fig:teaser}). \emph{Within a trajectory}, it can bypass inefficient detours within a demonstration, allowing a policy to learn shorter routes to success. \emph{Across successful trajectories}, it can combine complementary behaviors and expose continuations that were never recorded together. \emph{From failure trajectories}, it can recover useful behavior from failed episodes by connecting states along an unsuccessful execution to a successful continuation. 

Despite this promise, realizing this in real-world robot manipulation is difficult. A stitch must be both \emph{useful}, offering a better continuation, and \emph{feasible}, with actions that physically reach its target. Most prior methods that stitch trajectories typically assume access to low-dimensional or privileged state~\citep{yu2025astro,lee2025scots,hepburn2024stitchingbc,li2024diffstitch,char2021bats}, while methods designed to improve mixed-quality robot data typically filter or reweight recorded behavior without supplying the actions needed to connect desirable segments~\citep{du2023behaviorretrieval,hejna2025deminf,agia2025cupid,dass2026datamil}. We ask if we can get the best of both.

We introduce \textbf{N}ew \textbf{E}dges from \textbf{E}xisting \textbf{D}emonstrations through \textbf{L}ocal \textbf{E}diting (\method{}), which identifies valuable segments with no existing connection, and proposes verified action bridges to connect them. Our key insight is that \textbf{rather than learning a globally accurate dynamics model or task-specific value function and asking either to reason over long horizons, we can reduce stitching in high-dimensional spaces to a series of locally supervised reachability problems.} 

Specifically, the feasibility and duration of a local connection can be learned directly from recorded action sequences. Across any episode, successful, suboptimal, or failed, each short segment is an example of which two observations can be connected, the specific action sequence that connects them, and the exact number of steps it takes. We use this insight to train two modules. Given source and target observations, a goal-conditioned diffusion Inverse Dynamics Model (IDM) generates action sequences that attempt to reach from source to target. In parallel, we supervise an action-conditioned verifier that receives the same source and target observations and predicts whether, and after how many steps, the proposed actions reach the target. 


To make these connections useful for policy learning, \method{} turns accepted bridges into action sequences at recorded observations. Pairing each bridge with the originally-recorded actions leading up to its source exposes the synthesized actions from multiple preceding observations, without needing to generate intermediate images. Original training examples remain available, including those the bridge bypasses, so the policy receives new ways to make progress without losing supervision for the situations and behaviors covered by the original demonstrations.

Our contributions are threefold: 

\textbf{(1) An offline stitching method that finds useful connections and verifies bridges with an action-conditioned reachability model.} 

\textbf{(2) A training formulation that adds bridge actions while preserving recorded supervision, without synthesizing observations or requiring a shared coordinate frame.}

\textbf{(3) Evaluations on a real robot and in simulation of success, completion efficiency, recovery, and key design decisions.}

\vspace{-1.5em}
\section{Related Work}\label{sec:related_work}
\vspace{-1em}

\textbf{Explicit Stitching of Offline Datasets.}
Explicit dataset stitching materializes a connection between recorded segments of a trajectory as new training data. Existing methods select connections using signals such as task value, return, temporal structure, or coverage~\citep{char2021bats,hepburn2024stitchingbc,li2024diffstitch,lee2025scots,yu2025astro,chen2026extendable}. They assess connectivity through learned dynamics, successor likelihoods, or proximity in a learned temporal metric~\citep{char2021bats,hepburn2024stitchingbc,li2024diffstitch,lee2025scots,yu2025astro}. Generative approaches additionally synthesize states or longer trajectories for policy training~\citep{li2024diffstitch,zhou2024bellman,lee2025scots,yu2025astro,chen2026extendable}. \method{} instead formulates multistep connectivity as action-conditioned reachability between recorded observations, adding action supervision without reconstructing intermediate images. Connection utility depends on continuation lengths and success rather than a learned value function.

\textbf{Curation via Filtering, Retrieval, and Reweighting.}
Filtering, retrieval, and reweighting methods determine which recorded behavior should influence policy learning~\citep{du2023behaviorretrieval,wu2024ssdf,hejna2025deminf,chen2025demoscore,agia2025cupid,dass2026datamil}. They vary in granularity, from complete demonstrations to subtrajectories or individual transitions, and use signals such as relevance, progress, graph-based value, or backward reachability to identify useful behavior~\citep{memmel2025strap,yin2024gsr,yue2024diverse,wang2025sbr,chen2026sarm}. These approaches can recover useful supervision from heterogeneous data while leaving the recorded action targets unchanged. More broadly, reusing recorded trajectories to construct corrective supervision has been explored in learned models~\citep{venkatraman2015improving} and
policies~\citep{ke2024ccil,zhang2024dmd}. \method{} similarly creates supervision from existing trajectories, but constructs alternative action sequences connecting recorded observations, rather than corrections for predicted states.

\textbf{Implicit Stitching and Compositional Planning.} 
Offline RL can improve on recorded behavior by propagating value across transitions and composing favorable continuations through policy learning~\citep{kumar2022should,singh2021chaining,chebotar2023qtransformer,park2023hiql,hansenestruch2023idql}. Goal-conditioned formulations also relate action selection to learned estimates of future reachability~\citep{eysenbach2022contrastive,kaelbling1993learning}. Generative planners construct trajectories or subgoals at inference~\citep{janner2022diffuser,ajay2023decision,black2024susie,huang2024subgoal,kim2024ssd,luo2025compdiffuser}, while graph-based approaches search over recorded states using learned goal-reaching relationships~\citep{eysenbach2019sorb,emmons2020sgm}. Related policy-steering methods search over or evaluate action proposals using predicted outcomes at deployment~\citep{jain2025sailor,wu2025forewarn,ren2024hybrid}. In these approaches, composition is expressed through learned values and policies, or inference-time planning and steering. \method{} materializes its connections offline as explicit action supervision, so the resulting imitation policy requires neither connection search nor verification at deployment. We select baselines from each category to evaluate against in Section~\ref{sec:exp_overview}.

\section{How \method{}work Works}
\label{sec:method}
\vspace{-2mm}

\method{} answers two separate questions when rewriting a dataset: whether a stitch is \emph{useful}, and whether a stitch is \emph{feasible}. 
Candidate selection addresses the first question. An Inverse Dynamics Model (IDM) and an action-conditioned verifier address the second. 
We formalize the notation in \S\ref{sec:problem_formulation}, the selection of useful connections in \S\ref{sec:stitching_algorithm}, action generation and verification in \S\ref{sec:training_idm_verifier}, and policy training on the rewritten dataset in \S\ref{sec:dataset_construction}. Additional details in Appendix~\ref{app:method_details}.
\vspace{-2mm}

\begin{figure}[t]
    \centering
    \includegraphics[width=\linewidth]{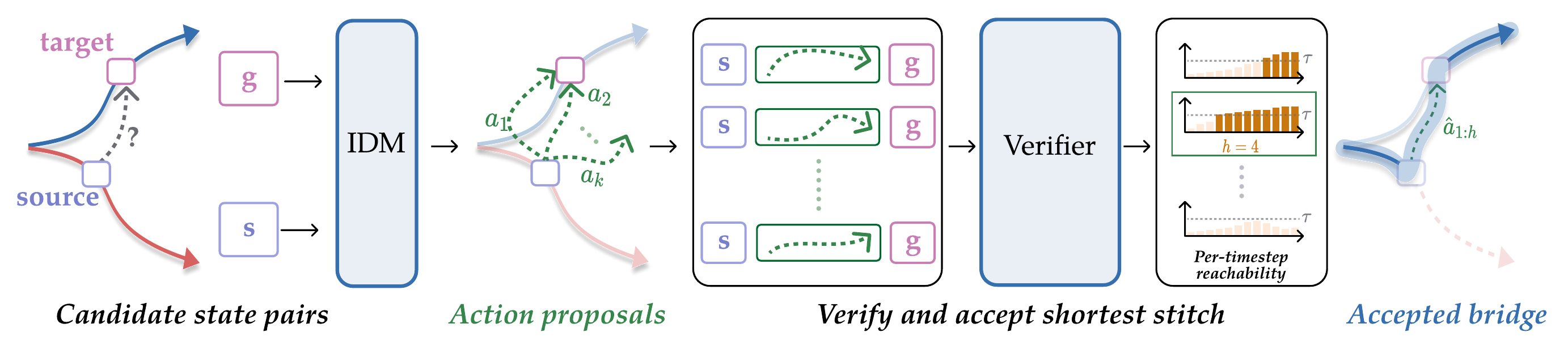}
    \caption{\textbf{From candidate connection to verified bridge.}
    Given a source-target pair $(s,g)$, the inverse dynamics model (IDM) samples $K$ action proposals. The action-conditioned verifier predicts whether and when each proposal reaches $g$. We select the shortest accepted  prefix as a bridge.}
    \label{fig:stitching_pipeline}
    \vspace{-4mm}
\end{figure}

\subsection{Notation}
\label{sec:problem_formulation}
\vspace{-2mm}
Let $\mathcal D=\{(\tau^i, y_i)\}_{i=1:N}$ be a dataset of $N$ trajectories $\tau^i=(o^i_1,a^i_1,\ldots,a^i_{T_i-1},o^i_{T_i}),$ where $y_i$ indicates whether trajectory $i$ succeeds or fails. Each observation $o_t^i=(x_t^i,p_t^i)$ contains multi-view RGB images $x_t^i$ and robot proprioception $p_t^i$. We use $i,j$ to index trajectories and $t,t'$ to index timesteps within them. 
A stitch connects a source observation $s=o_t^i$ to a target observation $g=o_{t'}^j$ through a newly generated action sequence. We use \emph{proposal} for a length-$H$ action sequence $\hat{a}_{1:H}$ sampled from the IDM, \emph{bridge} for an accepted prefix $b=(s,g,\hat a_{1:\ell})$, where $\ell\leq H$, and \emph{stitching} for the operation that incorporates a bridge into the policy-training data.

\subsection{Where should we stitch?}
\label{sec:stitching_algorithm}
\vspace{-2mm}
A feasible connection is not necessarily useful. Connecting nearby observations may simply reproduce existing behavior, or worse, introduce an unnecessary detour. We therefore search for source-target pairs whose recorded continuations offer an opportunity to improve the dataset.

We define three sources of candidate pairs, with different notions of ``improvement'' as illustrated in Figure~\ref{fig:teaser}, and retrieve candidates from these categories before proposing actions.
A \emph{within-trajectory shortcut} (S1) bypasses a longer segment of a successful trajectory.
A \emph{cross-trajectory transplant} (S2) connects a successful source to a shorter successful continuation in another trajectory.
A \emph{failure funnel} (S3) connects a source on an unsuccessful trajectory to an observation on a successful trajectory. Appendix~\ref{app:stage_ablation} isolates the effect when training on different subsets of these three stages.

\textbf{Defining a good stitch.}
For a source $s=o_t^i$ and target $g=o_{t'}^j$ on successful trajectories, the recorded source continuation requires $T_i-t$ steps, while the target has $T_j-t'$ steps remaining. If an $\ell$-step bridge reaches $g$, its potential reduction in completion steps is 
\begin{equation}
\Delta_{\mathrm{cont}}(b) = (T_i-t) -\left[\ell+(T_j-t')\right].
\label{eq:continuation_cost}
\end{equation}
We retain successful-to-successful bridges only when $\Delta_{\mathrm{cont}}(b)>0$. 
This criterion favors connections whose shorter remaining path outweighs the cost of reaching the target, conditional on the bridge succeeding and the target's continuation remaining applicable.

For failed sources, there is no successful continuation against which to measure steps saved. We instead seek connections to successful trajectories, preferring the shortest combined bridge and remaining successful continuation. These bridges introduce action supervision toward success from states that would otherwise contribute no successful continuation.%
\footnote{Enumerating all source-target pairs would be expensive. In practice, we use the IDM's visual encoder features to retrieve a finite neighborhood of candidate targets for each source. Retrieval identifies promising endpoints but does not establish that they can be connected.}

We retain at most one bridge from each source, selecting the accepted bridge with the shortest combined bridge and successful suffix, $\ell+(T_j-t')$. For successful sources, minimizing this quantity also maximizes the estimated continuation gain $\Delta_{\mathrm{cont}}$, and we require that gain to be positive. The same bridge-plus-suffix criterion ranks failure funnels. We discuss how to leverage these bridges for training in the following Section~\ref{sec:dataset_construction}.

\subsection{Training the needle and thimble}
\label{sec:training_idm_verifier}
\vspace{-2mm}
A useful connection improves the dataset only if its proposed actions reach the intended target. We address this problem by individually training an IDM and verifier, thus separating the challenges of action generation and verification~\citep{swamy2026likelihood}. At a local timescale, \textbf{each segment provides examples of \textit{which} observations can be connected, and \textit{how}}. We leverage this insight to design positive supervision from short-horizon segments within both successful and failure trajectories, thus expanding the amount of usable training data for our IDM and verifier.

\textbf{Goal-conditioned action proposals.} Following the goal-conditioned inverse-model formulation in~\citet{pathak2018zeroshot}, we instantiate the IDM as a goal-conditioned diffusion model
\begin{equation}
    \hat{a}_{1:H} \sim \pi_{\theta}\left(\cdot \mid s, g \right),
    \label{eq:idm}
\end{equation}
trained with the standard diffusion denoising objective~\citep{ho2020denoising,chi2023diffusionpolicy}. Training examples come from demonstrated segments known to be reachable within the action horizon $H$. For a source $s=o_t^i$, we sample an offset $1\leq\delta\leq H$, set $g=o_{t+\delta}^i$, and supervise the model with recorded actions up to that goal. If $\delta < H$, we hold the last action at the goal pose to fill the action horizon, thereby also encouraging greater sensitivity to the specific action that reaches the goal. At stitching time, the trained IDM samples $K$ action proposals for each candidate pair.

\textbf{Action-conditioned verification.} 
A goal-conditioned IDM produces plausible actions; the verifier needs to certify \emph{whether} the goal is reached by this set of actions, and \emph{how early}.%
\footnote{A ``thimble'', if you will, preventing harm from bad stitches.}
Thus, the verifier receives a source, goal, and proposed action sequence and predicts whether the goal has been reached by each prefix:
\begin{equation}
(r_1,\ldots,r_H)=f_\phi(s,g,a_{1:H}),
\label{eq}
\end{equation}
where $r_h$ scores whether executing $a_{1:h}$ from $s$ reaches $g$ at or before step $h$. 
Causal attention over the action tokens prevents $r_h$ from depending on actions after step $h$. The verifier and IDM share a visual encoder for the source and goal observations, and can be trained in parallel.

Predicting reachability at multiple action-prefix lengths lets us select both a connection and its duration, which is needed to evaluate its contribution to steps to completion ($\ell$ in Eq~\ref{eq:continuation_cost}). For recorded short segments, the actions and their designated target observations provide direct supervision for local reachability. We can therefore assess whether an action sequence connects two observations without estimating long-horizon returns or reconstructing intermediate states.

We construct three categories with the same anchor for training. Formally, let $o_{t}^i$ be the shared source observation, and $a_{t:{t+\delta^+}}$ the recorded action chunk from $o_{t}^i$, with $\delta^+\leq H$. Let $A_t^i$ denote this chunk padded to length $H$ by holding the goal pose when $\delta^+<H$, as described for the IDM above.

We sample a single positive \emph{within-horizon goal} $g^+=o_{t+\delta^+}^i$, whose binary target is positive for every $h\geq\delta^+$ and negative beforehand.
We also pair with a \emph{beyond-horizon goal} $g^{\mathrm{far}}=o_{t+\delta^{\mathrm{far}}}^i$, where $\delta^{\mathrm{far}}>H$, and with a \emph{cross-trajectory goal} $g^\times=o_{t'}^j$, where $j\neq i$. 
These latter two categories are negative pairs to the first positive anchor. Note that all pairs are conditioned on the same action chunk $A_t^i$. In particular, labeling our cross-trajectory tuple $\left(o_{t}^i, g^\times, A_t^i\right)$ negative does not mean that $\left(o_{t}^i, g^\times, A'\right)$ is negative for every possible $A'$. In fact, finding an alternative action chunk that connects such cross-trajectory observations is the objective of stitching.

We train all prefix outputs with binary cross-entropy. Since false-positive predictions introduce incorrect action supervision, we select a threshold $\tau_h$ for each horizon on held-out trajectories, prioritizing a low empirical false-positive rate~\citep{tong2018neyman}. We consider a prefix supported when its score exceeds the threshold used for its stitch stage.

\begin{figure}[t]
    \centering
    \includegraphics[width=\linewidth]{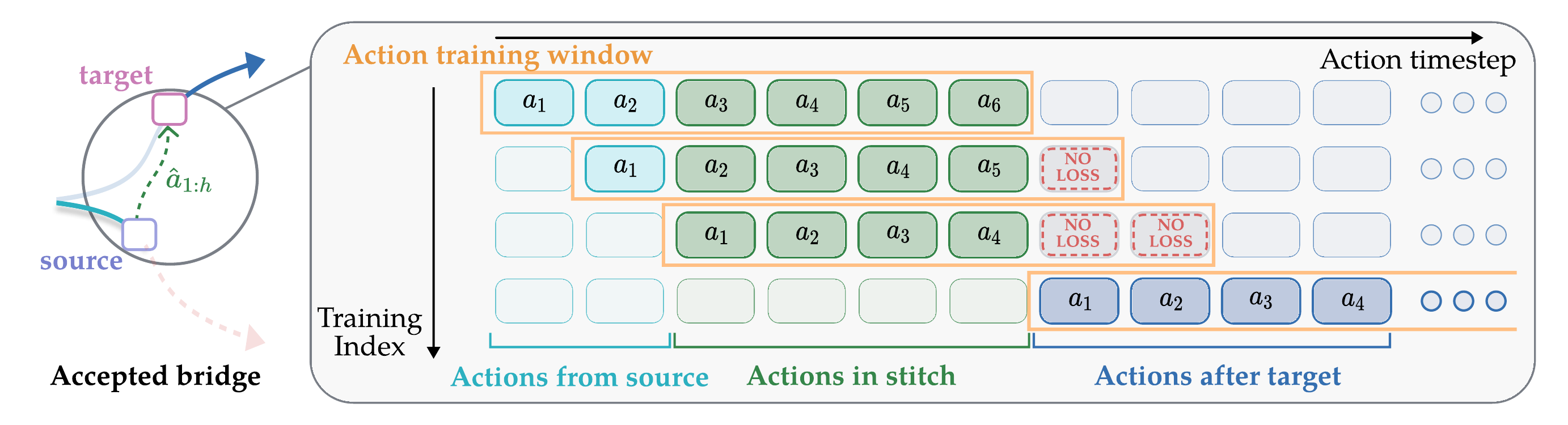}
    \caption{\textbf{Training on a bridge without intermediate observations.}
    Each example pairs a recorded observation with the original actions leading to the bridge (cyan), followed by the generated bridge actions (green). The loss stops when the bridge ends (red). Original demonstration windows remain available as separate training examples (blue).}
    \label{fig:window_construction}
    \vspace{-4mm}
\end{figure}

\subsection{Training a Policy from Stitches}
\label{sec:dataset_construction}
\vspace{-2mm}

Each policy-training example pairs recorded observations with a short sequence of actions to learn. A bridge supplies new actions but no images along the way. Its endpoint may also differ slightly from the recorded target, and the target demonstration may express its actions in a different coordinate frame, as in UMI~\citep{chi2024universal,umicommunity2026,patel2026behavior}. We account for all three limitations when constructing training examples (Figure~\ref{fig:window_construction}).

Consider an accepted bridge $b=(o_t^i,g,\hat a_{1:\ell})$. Rather than constructing a single training example at its source, we also construct examples beginning at preceding recorded observations. For a window starting $q$ steps before the bridge, the action target is

\vspace{-2mm}
\begin{equation}
    A^{(q)}
    =
    \left(
    a_{t-q}^i,\ldots,a_{t-1}^i,
    \hat a_1,\ldots,\hat a_\ell
    \right).
    \label{eq:bridge_window}
\end{equation}
Varying the starting point exposes the same connection through multiple training windows, allowing the policy to learn the new continuation from several preceding observations. Further, if a bridge ends before the policy's fixed action prediction horizon, we mask all action positions after its endpoint to prevent the policy from being trained on an artificial concatenation of the bridge and the target demonstration. Even when a bridge approximately reaches its target, the recorded actions following that target may not be valid from the bridge's actual endpoint. We instead learn the target continuation separately, paired with its own recorded observations.

\textbf{Preserving original coverage.}
Because a stitch creates an alternative local continuation rather than an exact synthetic trajectory, we still keep the original actions at bridge sources and along the segments they bypass, and control their sampling frequency relative to the new bridge windows. Our training construction preserves the original supervision and stops the new action target at the bridge endpoint. App.~\ref{app:sampling_ablation} shows that removing
these safeguards generally reduces downstream success.
\section{Experiments}
\label{sec:exp_overview}

\begin{figure*}[t]
    \centering
    \includegraphics[width=\textwidth]{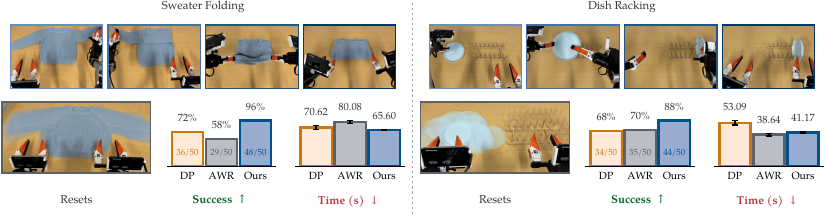}
    \caption{\textbf{Real-robot tasks and evaluation.} We evaluate \method{} on two real-robot bimanual tasks: Sweater Folding and Dish Racking. We show task sequences, reset distributions, success rates, and completion times. Each policy is evaluated 50 times per task at paired starting configurations across all methods. Completion times show mean and standard error over successful trials.}
    \label{fig:real_robot_results}
    \label{fig:real_tasks}
    \vspace{-4mm}
\end{figure*}

We evaluate whether rewriting an offline robot dataset with \method{} improves downstream policy performance, and, more importantly, \emph{how} the rewritten data changes the behavior learned by the policy. After describing the experimental setup (\S\ref{sec:exp_setup}), we first measure task success on real-robot and simulated manipulation tasks (\S\ref{sec:exp_succcess}). Then, we measure completion within a fixed budget on both real-robot and simulated tasks, and use controlled simulation experiments to isolate two effects of stitching: reaching useful states sooner and extending supervision to states encountered in failed trajectories (\S\ref{sec:exp_why_stitches}). Finally, we test whether verification is necessary for selecting useful bridges (\S\ref{sec:exp_verifier}) and examine \method{}'s performance when the demonstrations are already near-optimal (\S\ref{sec:exp_dexmimicgen}).

\subsection{Experimental Setup}
\label{sec:exp_setup}
\vspace{-2mm}

\textbf{Tasks and data.}
We evaluate sweater folding and dish racking on a real robot, using demonstrations collected with handheld UMI devices~\citep{chi2024universal,patel2026behavior}; policies receive wrist-camera RGB and proprioception measured relative to each trajectory's starting position (Figure~\ref{fig:real_robot_results}). In simulation, we use the suboptimal Robomimic Multi-Human \emph{Can}, \emph{Square}, and \emph{Transport} tasks~\citep{mandlekar2022robomimic}, with the same 40 successful demonstrations for every method and 150 additional failures for methods that use them. We leverage simulation because it permits controlled resets that let us compare continuations from identical states. We also test three DexMimicGen tasks with near-optimal demonstrations~\citep{jiang2025dexmimicgen}. All methods use RGB, proprioception, and episode outcomes, without per-step rewards, privileged object state, or new environment interaction.

\textbf{Baselines.}
We compare against an explicit stitching baseline \textbf{MBTS}~\citep{hepburn2024stitchingbc}, selection and reweighting baselines \textbf{SBR}~\citep{wang2025sbr}, \textbf{AWR}~\citep{peng2019awr,kostrikov2022iql}, and \textbf{SARM}~\citep{chen2026sarm}, and an implicit stitching baseline \textbf{IDQL}~\citep{hansenestruch2023idql}. We adapt SARM to the same level of supervision as other baselines, without the annotated subtask boundaries or language labels from its released implementation. IDQL reranks diffusion-policy samples with the IQL critic also used by AWR. Across tasks, all methods use the same DINOv3 vision encoder~\citep{simeoni2025dinov3} and Diffusion Policy architecture~\citep{chi2023diffusionpolicy}, so comparisons isolate differences in training supervision. App.~\ref{app:datasets} provides more information on evaluation details and task setups; App.~\ref{app:baseline_tuning} shows more detail on baseline and our tuning efforts. 

\subsection{Does \method{} Improve Policy Success?}
\label{sec:exp_succcess}
\vspace{-2mm}

\begin{wrapfigure}{R}{0.48\textwidth}
    \vspace{-4mm}
    \centering
    \includegraphics[width=\linewidth]{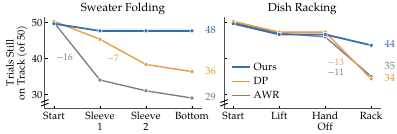}
    \caption{\footnotesize \textbf{Fewer failures across task stages.} Trials completing each successive stage, out of 50 per policy and task, from video review. Endpoint labels give final success counts, and the negative numbers mark each baseline's largest loss of trials at a single stage.}
    \label{fig:real_robot_stage_progress}
    \vspace{-4mm}
\end{wrapfigure}

\textbf{Rewriting improves real-robot success.} \method{} improves success by 24 percentage points on sweater folding and 18 points on dish racking over the strongest baseline (Figure~\ref{fig:real_robot_results}). We examine which stages account for these gains by reviewing all 50 trials per policy and task.

We seek to localize where and how imperfect demonstrations affect baselines. In Figure~\ref{fig:real_robot_stage_progress}, we decompose where each checkpoint fails over the course of a rollout. For both baseline policies, their main failure mode is repeating a suboptimal motion from demonstrations cyclically, such as folding and unfolding a sleeve, or circling its gripper around the rack for prolonged time before placing imprecisely. Appendix~\ref{app:real_robot_retracing} complements this analysis, showing \method{} having the lowest percentage of cyclic motion repeats across the two tasks. 

\begin{figure*}[t]
    \centering
    \includegraphics[width=\textwidth]{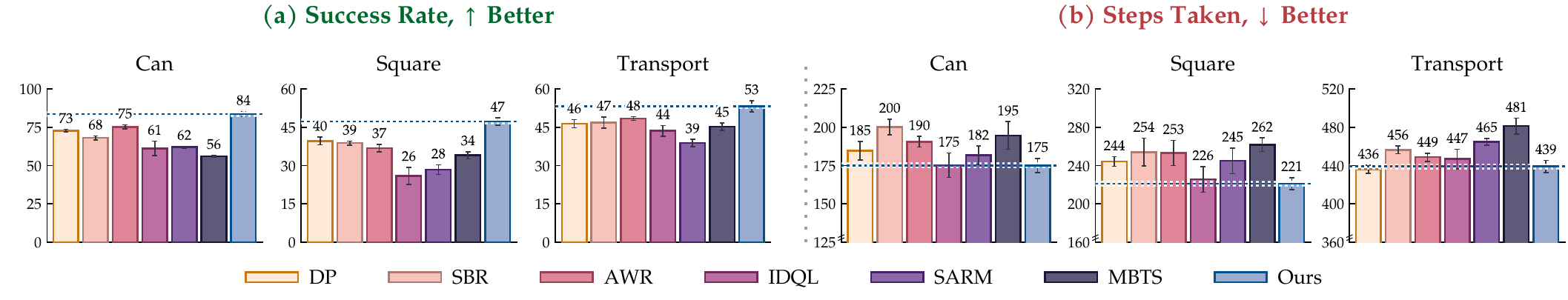}
    \caption{\textbf{Policy performance on suboptimal Robomimic demonstrations.} \textbf{(a)} \method{} achieves higher average success than all baselines. \textbf{(b)} Its successful rollouts take fewer steps on average than DP's on \textit{Can} and \textit{Square}. Bars show the mean and standard error over five training seeds, with 50 held-out initial conditions per seed; completion steps are averaged over successful trials.}
    \label{fig:robomimic_main_results}
    \vspace{-3mm}
\end{figure*}

\textbf{The same advantage appears with suboptimal simulated demonstrations.}
On the three Robomimic tasks, \method{} achieves higher average success than every evaluated baseline (Figure \ref{fig:robomimic_main_results}). Averaged across tasks, rewriting improves success by 8.4 percentage points over standard DP and by 6.9 points over the strongest competing baseline on each task. Though these aggregate success rates establish that rewriting helps, they do not by themselves explain why. To identify which parts of the design contribute to these gains, Appendix~\ref{app:stage_ablation} isolates the stitching stages, Appendix~\ref{app:sampling_ablation} ablates different training variations with the stitched dataset, and Appendix~\ref{app:bridge_ablation} tests whether the gain comes from the bridge actions themselves. We next use the controlled resets available in simulation to examine what the accepted stitches actually change. 

\subsection{What does \method{} change about the learned behavior?}
\label{sec:exp_why_stitches}
\vspace{-2mm}

We first measure whether rewriting improves completion within a fixed budget on real-robot and simulated tasks. We then use controlled simulation resets to test whether accepted bridges shorten the remaining path to success and whether failure funnels improve recovery from states outside the successful-demonstration distribution.

\begin{wrapfigure}{R}{0.48\textwidth}
    \vspace{-4mm}
    \centering
    \includegraphics[width=\linewidth]{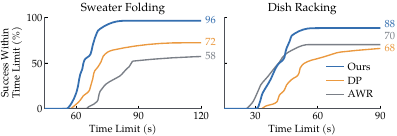}
    \caption{\footnotesize \textbf{\method{} finishes more trials within a fixed budget.} Labels give final success rates per time limit over the 50 paired trials.}
    \label{fig:real_robot_completion_budget}
    \vspace{-4mm}
\end{wrapfigure}

\begin{figure*}[t]
    \centering
    \includegraphics[width=0.99\textwidth]{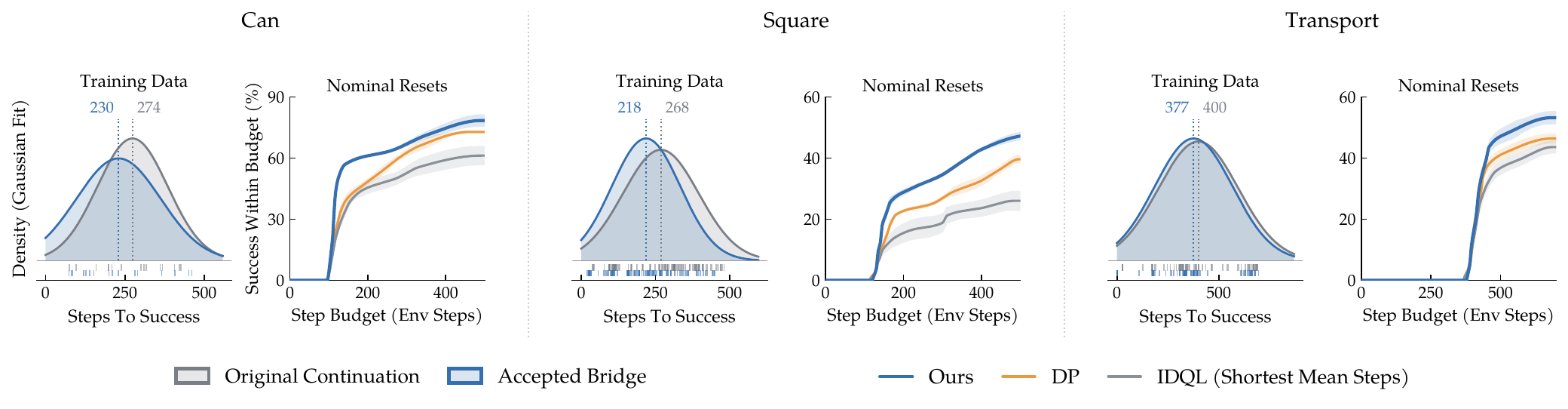}
    \vspace{-2mm}
    \caption{\textbf{Shorter continuations and higher success within a step budget.} For each task, the left panel shows remaining completion steps after executing either the original continuation or an accepted bridge, restricted to pairs where both continuations succeed. On average, bridges lead to shorter remaining continuations. The right panel shows the fraction of matched nominal starts solved within each budget. \method{} reaches comparable success sooner and solves more starts by the full horizon. We plot mean and S.E. for five training seeds, each with 50 held-out starts.}
    \label{fig:stitching_completion}
    \vspace{-4mm}
\end{figure*}

\textbf{Rewriting yields more success within a fixed budget.} Mean completion time among successes can be misleading: a policy that solves fewer difficult starts may appear faster. We therefore measure the fraction of all evaluation trials completed within different time-budget thresholds, with failures counted against it, shown in Figure~\ref{fig:real_robot_completion_budget}. On both real-robot tasks, \method{} reaches comparable success sooner and ultimately completes more trials than either baseline. The same pattern holds in simulation, shown in Figure~\ref{fig:stitching_completion}, right. Evaluated from matched initial conditions, \method{} reaches comparable success in fewer steps and solves more starts by the full horizon.

\textbf{Accepted bridges move the policy toward shorter suffixes.}
From each state that \method{} finds a valid stitch for, we execute the stitch's full action sequence and the original continuation for the same number of steps, then resume the same DP from the two resulting states in simulation. This can be viewed as an advantage-style comparison under DP: the two continuations begin from the same source, and we ask whether our stitches leave the policy in a state from which success is quicker to achieve. The left panels of Figure~\ref{fig:stitching_completion} show that, across all three tasks, accepted bridges shift the distribution toward fewer remaining steps to success, confirming the behavior that the continuation-gain criterion from Equation~\ref{eq:continuation_cost} seeks to induce. 

\begin{figure*}[t]
    \centering
    \includegraphics[width=\textwidth]{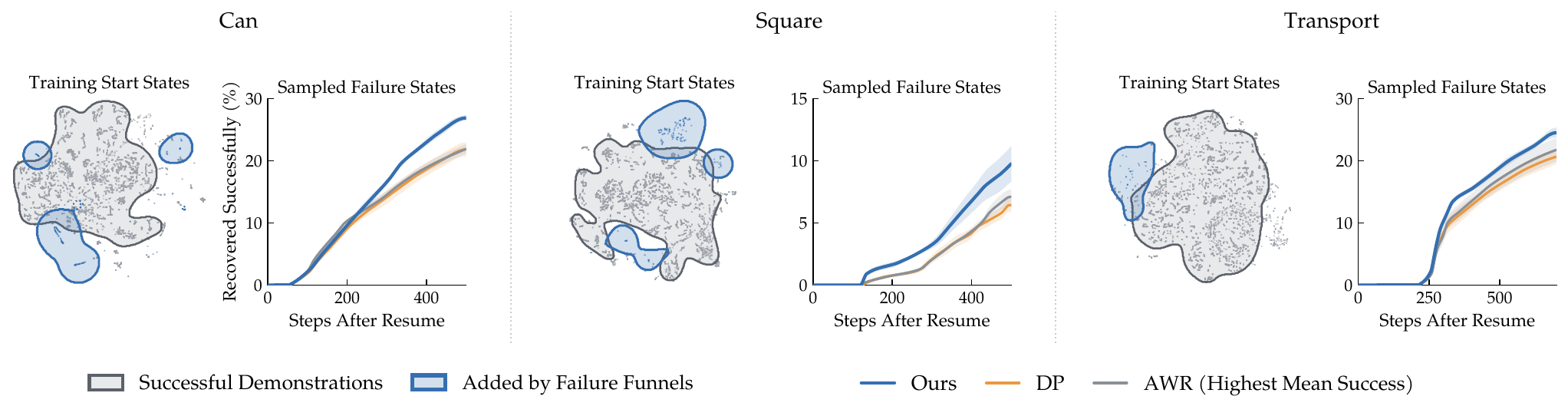}
    \caption{\textbf{Expanded training coverage and recovery from failure states.} For each task, the left panel shows successful-demonstration states and starts added by failure funnels, visualized as 90\% density regions in a t-SNE embedding of the visual encoder. The right panel measures recovery after resuming from OOD states, where \method{} recovers more often than DP and AWR, the non-DP baseline with the highest average success rate across all three tasks on nominal resets. Curves show mean and standard error for five training seeds, each with 50 held-out starts.}
    \label{fig:stitching_recovery}
    \vspace{-3mm}
\end{figure*}

\textbf{Failure funnels add supervision outside the successful-demonstration distribution.}
The second effect is qualitatively different. A failed trajectory may contain useful intermediate behavior even though it never reaches the task goal. A failure funnel connects such a state to a successful continuation, converting a state that previously had no successful action target into one with explicit supervision toward success. The left panels of Figure~\ref{fig:stitching_recovery} visualize this effect. Starts introduced by failure funnels occupy regions of the learned visual representation that extend beyond those covered by successful demonstrations. 

\textbf{Additional coverage from failure improves recovery.}
We test whether this broader supervision changes closed-loop behavior by resetting each policy to the same randomly-sampled failure states and measuring subsequent recovery. \method{} recovers more often than DP on all three evaluated populations (right panels of Figure~\ref{fig:stitching_recovery}) and also outperforms the strongest non-DP baseline, AWR.

\takeaway{Verified bridges connect recorded states to routes requiring fewer remaining steps to success. Connecting states from failed demonstrations to successful routes adds action supervision beyond the successful demonstrations and improves recovery.}

\subsection{Does verifying feasibility matter?}
\label{sec:exp_verifier}
\vspace{-2mm}
A useful connection must meet two separate tests. It must be useful: following the target's recorded actions should give the policy a better route to task completion. It must also be feasible: the proposed bridge actions must reach the target. Candidate selection estimates utility from recorded continuations; here we test whether the verifier predicts feasibility from the proposed actions, whether that prediction reflects physical bridge quality, and whether filtering on it improves downstream learning.

\begin{figure*}[t]
    \centering
    \includegraphics[width=0.98\textwidth]{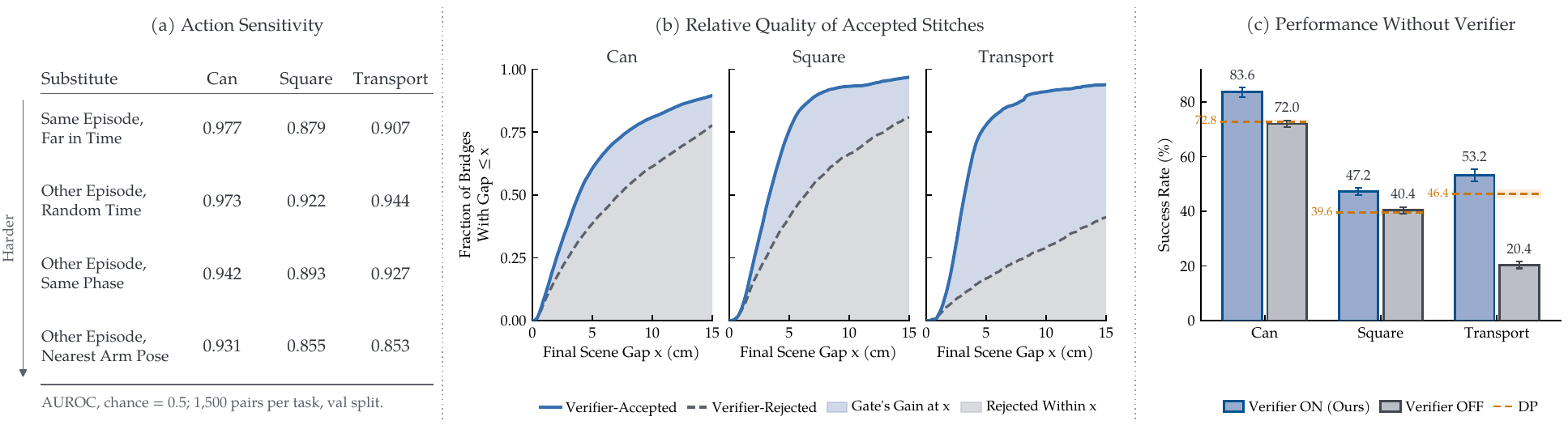}
    \caption{\textbf{Verifier-based selection connects action sensitivity, bridge quality, and policy success.} \textbf{(a)} With the source and target fixed, the verifier distinguishes demonstrated actions from logged substitutes on 1,500 held-out pairs per task. \textbf{(b)} Accepted proposals more often land near the target than rejected proposals. Scene gap is the larger of the arm and object position errors, measured only for this diagnostic; \method{} itself never observes object positions. \textbf{(c)} Removing verifier filtering reduces downstream success on all three tasks with the IDM and bridge counts per stage held fixed. Bars show success and standard error over five training seeds of 50 evaluations each.}
    \label{fig:verifier_analysis}
    \vspace{-3mm}
\end{figure*}

\textbf{The verifier distinguishes actions and selects closer landings.}
We first ask how sensitive the verifier is to the proposed action, as opposed to just relying on the source and goal inputs. For each held-out source--goal pair, we replace the demonstrated action chunk with a logged chunk from a distant time in the same episode or from another episode, selected randomly, at the same relative episode phase, or by nearest starting arm pose. Figure~\ref{fig:verifier_analysis}(a) shows that the verifier distinguishes the demonstrated actions across all substitute types (AUROC 0.85--0.98), including the closest-pose substitutes on each task. 

Next, we replay proposed bridges in simulation and measure the larger of the arm and object position errors relative to the target (Figure~\ref{fig:verifier_analysis}(b)); this error is available only in simulation and is never observed by \method{}. Accepted bridges are more likely than rejected bridges to finish within 5\,cm on all three tasks, by 21.9, 34.6, and 61.1 percentage points on \textit{Can}, \textit{Square}, and \textit{Transport}, respectively. In absolute terms, about 61\%, 76\%, and 78\% of accepted bridges land within 5 cm respectively on these three tasks. These results show that the verifier responds to the action sequence and selects bridges that more often land near the target.

\textbf{Verifier-based selection improves the downstream policy.}
To test whether this selection matters for learning, we compare performance against policies trained on just the IDM proposals, while keeping the number of training bridges per stage the same as what was accepted by the verifier. The resulting policy loses success on all three tasks (Figure~\ref{fig:verifier_analysis}(c)); on \textit{Transport}, success falls from 53.2\% to 20.4\%, a decrease of 32.8 percentage points (61.7\% relative). Thus, the added data's value depends on which bridges are selected, beyond the number of examples added.

\takeaway{The verifier selects bridges that more reliably reach their intended targets, where recorded behavior offers a shorter route to completion. Without this filtering, the same number of added bridges produces policies performing near or below standard behavior cloning.}

\subsection{What if existing demonstrations are already near-optimal?}
\label{sec:exp_dexmimicgen}
\vspace{-2mm}

\begin{wraptable}{R}{0.52\textwidth}
    \vspace{-1em}
    \centering
    \small
    \setlength{\tabcolsep}{7pt}
    \renewcommand{\arraystretch}{1.08}
    \begin{tabular}{lcc}
        \toprule
        Task & DP & \textbf{Ours} \\
        \midrule
        Coffee & 62.0\,\tabsem{1.2} & \textbf{63.3}\,\tabsem{0.7} \\
        Three Piece Assembly & 51.3\,\tabsem{0.7} & \textbf{55.3}\,\tabsem{1.8} \\
        Threading & \textbf{46.0}\,\tabsem{1.2} & \textbf{46.0}\,\tabsem{1.2} \\
        \bottomrule
    \end{tabular}
    \caption{\footnotesize \textbf{Policy success with near-optimal DexMimicGen demonstrations.} \method{} matches or modestly improves on DP across the three tasks. Values are success rates (\%), reported as mean and standard error over three training seeds with 50 held-out initial conditions per seed.}
    \vspace{-2em}
    \label{tab:dexmimicgen_success_best}
\end{wraptable}

Table~\ref{tab:dexmimicgen_success_best} reports how much rewriting helps when demonstrations are already near-optimal, and highlights an important boundary condition. \method{}'s percentage gain is smaller, at a maximum gain of four percentage points. Thus, \method{} is most valuable when an offline dataset contains complementary, inefficient, or unsuccessful behavior that can be recomposed into better supervision. When the demonstrations are already close to the desired behavior, rewriting has correspondingly less to add; nevertheless, it does not reduce success relative to standard BC.

\subsection{Limitations and Future Work}
\vspace{-2mm}

The smaller gains on near-optimal demonstrations suggest that \method{} depends on opportunities for useful recomposition; future work could identify these opportunities and guide data collection where coverage is insufficient. Although our experiments show that local bridges shorten continuations and improve recovery, connections requiring sustained contact, substantial object reconfiguration, or feedback during execution may exceed the short-bridge formulation. Future work could explore variable-length IDM predictions and additional recorded modalities such as force to better capture contact, while addressing cases that require feedback within a bridge. 
\vspace{-2mm}
\section{Conclusion}
\label{sec:conclusion}
\vspace{-2mm}

We present \method{}, an offline dataset-augmentation method for composing useful behavior within a dataset of imperfect robot demonstrations. \method{} reduces stitching to short, locally supervised reachability problems. Across real-robot and simulated manipulation tasks, our results show that this formulation can turn inefficient, complementary, and unsuccessful behavior into useful new supervision. More broadly, they suggest that improving offline robot data need not be limited to selecting which recorded behaviors to learn from; useful supervision can also be created by carefully connecting the behaviors already present.

\ificlrfinal
  \section{Acknowledgments}

Juntao Ren is supported by the Knight-Hennessy Fellowship and the NSF Graduate Research Fellowship Program (GRFP). 
This work was supported in part by the NSF Award \#2143601, \#2037101, and \#2132519, Toyota Research Institute, and Amazon. 
We would like to thank Gokul Swamy and all members of the REAL Lab at Stanford for their detailed feedback on paper drafts and figures, Austin Patel and Yihuai Gao for their help with real-robot setups, Mandi Zhao for her help with simulation setup, and Huy Ha for his advice in the early stages of this project. 
The views and conclusions contained herein are those of the authors and should not be interpreted as necessarily representing the official policies, either expressed or implied, of the sponsors.
\fi


\subsection*{AI use statement}
In this work, we used generative AI tools to integrate baselines from publicly released code into our training pipeline, support qualitative data analysis, and to obtain feedback on the manuscript, including critiques of experimental design, baseline choices, and analysis in the form of simulated peer review. The authors designed and ran all experiments and decided which feedback to act on. We have not used generative AI tools to generate synthetic datasets, develop theoretical models or conceptual frameworks, formulate mathematical claims or proofs, propose or refine hypotheses, assist with translation, or clean or reformat datasets. Additionally, we used generative AI tools to format the supplementary website, identify and summarize relevant literature, and format references. We reviewed all AI-assisted work, and all AI-suggested text edits were reviewed and revised by the authors. We take responsibility for the final content of this work, including text, claims, or artifacts produced with the aid of generative AI.

\subsection*{Reproducibility statement}
Appendix~\ref{app:datasets} documents dataset collection, composition, and selection, task setups, and checkpoint selection. Appendix~\ref{app:method_details} provides model architectures, optimization settings, hyperparameters, verifier calibration, stitching and training-window construction, and compute requirements. Appendix~\ref{app:baseline_tuning} details baseline adaptations and tuning. Evaluation and diagnostic protocols are described in Appendix~\ref{app:evaluation_details}. Additional results and ablations are provided in Appendix~\ref{app:additional_results}. We will open-source all the code and algorithms presented in this paper, from training to dataset rewriting.

\bibliography{iclr2027_conference}

@article{jain2025sailor,
  author  = {Arnav Kumar Jain and Vibhakar Mohta and Subin Kim and Atiksh Bhardwaj and Juntao Ren and Yunhai Feng and Sanjiban Choudhury and Gokul Swamy},
  title   = {A Smooth Sea Never Made a Skilled SAILOR: Robust Imitation via Learning to Search},
  journal = {CoRR},
  volume  = {abs/2506.05294},
  year    = {2025}
}

@inproceedings{du2023behaviorretrieval,
title     = {Behavior Retrieval: Few-Shot Imitation Learning by Querying Unlabeled Datasets},
author    = {Du, Maximilian and Nair, Suraj and Sadigh, Dorsa and Finn, Chelsea},
booktitle = {Proceedings of Robotics: Science and Systems},
year      = {2023},
doi       = {10.15607/RSS.2023.XIX.011}
}

@inproceedings{wu2024ssdf,
title     = {Learning From Imperfect Demonstrations With Self-Supervision for Robotic Manipulation},
author    = {Wu, Kun and Liu, Ning and Zhao, Zhen and Qiu, Di and Li, Jinming and Che, Zhengping and Xu, Zhiyuan and Tang, Jian},
booktitle = {2025 IEEE International Conference on Robotics and Automation (ICRA)},
pages     = {16899--16906},
year      = {2025},
doi       = {10.1109/ICRA55743.2025.11127918}
}

@inproceedings{hejna2025deminf,
title     = {Robot Data Curation with Mutual Information Estimators},
author    = {Hejna, Joey and Mirchandani, Suvir and Balakrishna, Ashwin and Xie, Annie and Wahid, Ayzaan and Tompson, Jonathan and Sanketi, Pannag R. and Shah, Dhruv and Devin, Coline Manon and Sadigh, Dorsa},
booktitle = {Proceedings of Robotics: Science and Systems},
address   = {Los Angeles, CA, USA},
month     = jun,
year      = {2025},
doi       = {10.15607/RSS.2025.XXI.023}
}

@inproceedings{agia2025cupid,
title     = {{CUPID}: Curating Data your Robot Loves with Influence Functions},
author    = {Agia, Christopher and Sinha, Rohan and Yang, Jingyun and Antonova, Rika and Pavone, Marco and Nishimura, Haruki and Itkina, Masha and Bohg, Jeannette},
booktitle = {Proceedings of The 9th Conference on Robot Learning},
series    = {Proceedings of Machine Learning Research},
volume    = {305},
pages     = {2907--2932},
year      = {2025}
}

@inproceedings{dass2026datamil,
  title = {{DataMIL}: Selecting Data for Robot Imitation Learning with Datamodels},
  author = {Dass, Shivin and Khaddaj, Alaa and Engstrom, Logan and Madry, Aleksander and Ilyas, Andrew and Martín-Martín, Roberto},
  booktitle = {The Fourteenth International Conference on Learning Representations},
  year = {2026},
  url = {https://proceedings.iclr.cc/paper_files/paper/2026/hash/033d9e8dbbbc0ad90e59222cf1db0fc2-Abstract-Conference.html}
}

@inproceedings{chen2025demoscore,
title     = {Curating Demonstrations using Online Experience},
author    = {Chen, Annie S. and Lessing, Alec M. and Liu, Yuejiang and Finn, Chelsea},
booktitle = {Proceedings of Robotics: Science and Systems},
year      = {2025},
doi       = {10.15607/RSS.2025.XXI.071}
}

@inproceedings{chi2023diffusionpolicy,
title     = {Diffusion Policy: Visuomotor Policy Learning via Action Diffusion},
author    = {Chi, Cheng and Feng, Siyuan and Du, Yilun and Xu, Zhenjia and Cousineau, Eric and Burchfiel, Benjamin C. M. and Song, Shuran},
booktitle = {Proceedings of Robotics: Science and Systems},
address   = {Daegu, Republic of Korea},
month     = jul,
year      = {2023},
doi       = {10.15607/RSS.2023.XIX.026}
}

@inproceedings{zhou2024bellman,
  title     = {Free from Bellman Completeness: Trajectory Stitching via Model-based Return-conditioned Supervised Learning},
  author    = {Zhou, Zhaoyi and Zhu, Chuning and Zhou, Runlong and Cui, Qiwen and Gupta, Abhishek and Du, Simon Shaolei},
  booktitle = {International Conference on Learning Representations},
  year      = {2024}
}

@inproceedings{janner2022diffuser,
  title     = {Planning with Diffusion for Flexible Behavior Synthesis},
  author    = {Janner, Michael and Du, Yilun and Tenenbaum, Joshua B. and Levine, Sergey},
  booktitle = {Proceedings of the 39th International Conference on Machine Learning},
  pages     = {9902--9915},
  volume    = {162},
  series    = {Proceedings of Machine Learning Research},
  publisher = {PMLR},
  year      = {2022}
}

@inproceedings{ajay2023decision,
  title     = {Is Conditional Generative Modeling All You Need for Decision-Making?},
  author    = {Ajay, Anurag and Du, Yilun and Gupta, Abhi and Tenenbaum, Joshua
               and Jaakkola, Tommi S. and Agrawal, Pulkit},
  booktitle = {International Conference on Learning Representations},
  year      = {2023}
}

@inproceedings{black2024susie,
  title     = {Zero-Shot Robotic Manipulation with Pretrained Image-Editing Diffusion Models},
  author    = {Black, Kevin and Nakamoto, Mitsuhiko and Atreya, Pranav and Walke, Homer and Finn, Chelsea and Kumar, Aviral and Levine, Sergey},
  booktitle = {International Conference on Learning Representations},
  year      = {2024}
}

@article{huang2024subgoal,
  title   = {Subgoal Diffuser: Coarse-to-fine Subgoal Generation to Guide Model Predictive Control for Robot Manipulation},
  author  = {Huang, Zixuan and Lin, Yating and Yang, Fan and Berenson, Dmitry},
  journal = {arXiv preprint arXiv:2403.13085},
  year    = {2024}
}

@article{hepburn2024stitchingbc,
  title   = {Model-based trajectory stitching for improved behavioural cloning and its applications},
  author  = {Hepburn, Charles A. and Montana, Giovanni},
  journal = {Machine Learning},
  volume  = {113},
  pages   = {647--674},
  year    = {2024},
  doi     = {10.1007/s10994-023-06392-z}
}

@inproceedings{li2024diffstitch,
  title     = {{DiffStitch}: Boosting Offline Reinforcement Learning with Diffusion-based Trajectory Stitching},
  author    = {Li, Guanghe and Shan, Yixiang and Zhu, Zhengbang and Long, Ting and Zhang, Weinan},
  booktitle = {Proceedings of the 41st International Conference on Machine Learning},
  pages     = {28597--28609},
  volume    = {235},
  series    = {Proceedings of Machine Learning Research},
  publisher = {PMLR},
  year      = {2024}
}

@inproceedings{kim2024ssd,
  title     = {Stitching Sub-Trajectories with Conditional Diffusion Model for Goal-Conditioned Offline RL},
  author    = {Kim, Sungyoon and Choi, Yunseon and Matsunaga, Daiki E. and Kim, Kee-Eung},
  booktitle = {Proceedings of the AAAI Conference on Artificial Intelligence},
  volume    = {38},
  pages     = {13160--13167},
  year      = {2024},
  doi       = {10.1609/aaai.v38i12.29215}
}

@inproceedings{lee2025scots,
  title     = {State-Covering Trajectory Stitching for Diffusion Planners},
  author    = {Lee, Kyowoon and Choi, Jaesik},
  booktitle = {Advances in Neural Information Processing Systems},
  volume    = {38},
  year      = {2025}
}

@inproceedings{chebotar2023qtransformer,
  title     = {{Q-Transformer}: Scalable Offline Reinforcement Learning via Autoregressive Q-Functions},
  author    = {Chebotar, Yevgen and Vuong, Quan and Hausman, Karol and Xia, Fei and Lu, Yao and Irpan, Alex and Kumar, Aviral and Yu, Tianhe and Herzog, Alexander and Pertsch, Karl and Gopalakrishnan, Keerthana and Ibarz, Julian and Nachum, Ofir and Sontakke, Sumedh Anand and Salazar, Grecia and Tran, Huong T. and Peralta, Jodilyn and Tan, Clayton and Manjunath, Deeksha and Singh, Jaspiar and Zitkovich, Brianna and Jackson, Tomas and Rao, Kanishka and Finn, Chelsea and Levine, Sergey},
  booktitle = {Proceedings of the 7th Conference on Robot Learning},
  pages     = {3909--3928},
  volume    = {229},
  series    = {Proceedings of Machine Learning Research},
  publisher = {PMLR},
  year      = {2023}
}

@inproceedings{park2023hiql,
  title     = {{HIQL}: Offline Goal-Conditioned RL with Latent States as Actions},
  author    = {Park, Seohong and Ghosh, Dibya and Eysenbach, Benjamin and Levine, Sergey},
  booktitle = {Advances in Neural Information Processing Systems},
  volume    = {36},
  year      = {2023}
}

@article{hansenestruch2023idql,
  title   = {{IDQL}: Implicit Q-Learning as an Actor-Critic Method with Diffusion Policies},
  author  = {Hansen-Estruch, Philippe and Kostrikov, Ilya and Janner, Michael and Grudzien Kuba, Jakub and Levine, Sergey},
  journal = {arXiv preprint arXiv:2304.10573},
  year    = {2023}
}

@inproceedings{wang2025sbr,
  title={Robust Offline Imitation Learning Through State-level Trajectory Stitching},
  author={Wang, Shuze and Mei, Yunpeng and Cao, Hongjie and Yuan, Yetian and Wang, Gang and Sun, Jian and Chen, Jie},
  booktitle={2025 IEEE/RSJ International Conference on Intelligent Robots and Systems (IROS)},
  pages={13497--13504},
  year={2025},
  organization={IEEE}
}

@inproceedings{eysenbach2022contrastive,
  title     = {Contrastive Learning as Goal-Conditioned Reinforcement Learning},
  author    = {Eysenbach, Benjamin and Zhang, Tianjun and Levine, Sergey and Salakhutdinov, Ruslan},
  booktitle = {Advances in Neural Information Processing Systems},
  volume    = {35},
  pages     = {35603--35620},
  year      = {2022}
}

@inproceedings{char2021bats,
  title     = {{BATS}: Best Action Trajectory Stitching},
  author    = {Char, Ian and Mehta, Viraj and Villaflor, Adam and
               Dolan, John M. and Schneider, Jeff},
  booktitle = {NeurIPS Workshop on Offline Reinforcement Learning},
  year      = {2021}
}

@inproceedings{yin2024gsr,
  title     = {Offline Imitation Learning Through Graph Search and Retrieval},
  author    = {Yin, Zhao-Heng and Abbeel, Pieter},
  booktitle = {Proceedings of Robotics: Science and Systems},
  year      = {2024}
}

@article{yu2025astro,
  title   = {{ASTRO}: Adaptive Stitching via Dynamics-Guided Trajectory Rollouts},
  author  = {Yu, Hang and Zhang, Di and Du, Qiwei and Zhao, Yanping and Zhang, Hai
             and Chen, Guang and Veas, Eduardo E. and Zhao, Junqiao},
  journal = {arXiv preprint arXiv:2511.23442},
  year    = {2025}
}

@inproceedings{eysenbach2019sorb,
  title     = {Search on the Replay Buffer: Bridging Planning and Reinforcement Learning},
  author    = {Eysenbach, Benjamin and Salakhutdinov, Ruslan and Levine, Sergey},
  booktitle = {Advances in Neural Information Processing Systems},
  volume    = {32},
  year      = {2019}
}

@inproceedings{emmons2020sgm,
  title     = {Sparse Graphical Memory for Robust Planning},
  author    = {Emmons, Scott and Jain, Ajay and Laskin, Michael and Kurutach, Thanard
               and Abbeel, Pieter and Pathak, Deepak},
  booktitle = {Advances in Neural Information Processing Systems},
  volume    = {33},
  pages     = {5251--5262},
  year      = {2020}
}

@inproceedings{luo2025compdiffuser,
  title     = {Generative Trajectory Stitching through Diffusion Composition},
  author    = {Luo, Yunhao and Mishra, Utkarsh A. and Du, Yilun and Xu, Danfei},
  booktitle = {Advances in Neural Information Processing Systems},
  volume    = {38},
  year      = {2025},
  note      = {Spotlight}
}

@inproceedings{chen2026extendable,
  title     = {Extendable Planning via Multiscale Diffusion},
  author    = {Chen, Chang and Hamed, Hany and Baek, Doojin and Kang, Taegu
               and Noh, Samyeul and Bengio, Yoshua and Ahn, Sungjin},
  booktitle = {Proceedings of the AAAI Conference on Artificial Intelligence},
  year      = {2026}
}

@inproceedings{memmel2025strap,
  title     = {{STRAP}: Robot Sub-Trajectory Retrieval for Augmented Policy Learning},
  author    = {Memmel, Marius and Berg, Jacob and Chen, Bingqing
               and Gupta, Abhishek and Francis, Jonathan},
  booktitle = {International Conference on Learning Representations},
  year      = {2025}
}

@inproceedings{chi2024universal,
title     = {Universal Manipulation Interface: In-The-Wild Robot Teaching Without In-The-Wild Robots},
author    = {Chi, Cheng and Xu, Zhenjia and Pan, Chuer and Cousineau, Eric and Burchfiel, Benjamin and Feng, Siyuan and Tedrake, Russ and Song, Shuran},
booktitle = {Proceedings of Robotics: Science and Systems},
address   = {Delft, Netherlands},
month     = jul,
year      = {2024},
doi       = {10.15607/RSS.2024.XX.045}
}

@misc{umicommunity2026,
  author       = {{UMI Robot Dataset Community}},
  title        = {{UMI Robot Dataset Community}},
  year         = {2026},
  howpublished = {\url{https://umi-data.github.io/}},
  note         = {Accessed July 2026}
}

@article{peng2019awr,
  title     = {Advantage-Weighted Regression: Simple and Scalable Off-Policy Reinforcement Learning},
  author    = {Peng, Xue Bin and Kumar, Aviral and Zhang, Grace and Levine, Sergey},
  journal   = {arXiv preprint arXiv:1910.00177},
  year      = {2019}
}

@inproceedings{kostrikov2022iql,
  title     = {Offline Reinforcement Learning with Implicit {Q}-Learning},
  author    = {Kostrikov, Ilya and Nair, Ashvin and Levine, Sergey},
  booktitle = {International Conference on Learning Representations},
  year      = {2022},
  url       = {https://openreview.net/forum?id=68n2s9ZJWF8}
}

@inproceedings{chen2026sarm,
  title     = {{SARM}: Stage-Aware Reward Modeling for Long Horizon Robot Manipulation},
  author    = {Chen, Qianzhong and Yu, Justin and Schwager, Mac and
               Abbeel, Pieter and Shentu, Fred and Wu, Philipp},
  booktitle = {The Fourteenth International Conference on Learning Representations},
  year      = {2026},
  url       = {https://proceedings.iclr.cc/paper_files/paper/2026/hash/1c92edb990a05f2269f0cc3afbb4c952-Abstract-Conference.html}
}

@inproceedings{mandlekar2022robomimic,
  title     = {What Matters in Learning from Offline Human Demonstrations for Robot Manipulation},
  author    = {Mandlekar, Ajay and Xu, Danfei and Wong, Josiah and
               Nasiriany, Soroush and Wang, Chen and Kulkarni, Rohun and
               Fei-Fei, Li and Savarese, Silvio and Zhu, Yuke and
               Mart{\'i}n-Mart{\'i}n, Roberto},
  booktitle = {Proceedings of the 5th Conference on Robot Learning},
  editor    = {Faust, Aleksandra and Hsu, David and Neumann, Gerhard},
  series    = {Proceedings of Machine Learning Research},
  volume    = {164},
  pages     = {1678--1690},
  publisher = {PMLR},
  year      = {2022},
  url       = {https://proceedings.mlr.press/v164/mandlekar22a.html}
}

@inproceedings{yue2024diverse,
  title     = {How to Leverage Diverse Demonstrations in Offline Imitation Learning},
  author    = {Yue, Sheng and Liu, Jiani and Hua, Xingyuan and Ren, Ju
               and Lin, Sen and Zhang, Junshan and Zhang, Yaoxue},
  booktitle = {Proceedings of the 41st International Conference on Machine Learning},
  series    = {Proceedings of Machine Learning Research},
  volume    = {235},
  pages     = {58037--58067},
  year      = {2024}
}

@inproceedings{jiang2025dexmimicgen,
  title     = {{DexMimicGen}: Automated Data Generation for Bimanual Dexterous Manipulation via Imitation Learning},
  author    = {Jiang, Zhenyu and Xie, Yuqi and Lin, Kevin and Xu, Zhenjia and Wan, Weikang and Mandlekar, Ajay and Fan, Linxi and Zhu, Yuke},
  booktitle = {2025 IEEE International Conference on Robotics and Automation (ICRA)},
  year      = {2025},
  publisher = {IEEE},
  doi       = {10.1109/ICRA55743.2025.11127809}
}

@article{patel2026behavior,
  title={Behavior Prompting Policy: Demonstrations as Prompts for Manipulation}, 
  author={Austin Patel and Ben Pekarek and Joel Enrique Castro Hernandez and Shuran Song},
  year={2026},
  journal={arXiv preprint arXiv:2606.30457},
  url={https://arxiv.org/abs/2606.30457}
}

@inproceedings{kaelbling1993learning,
  title = {Learning to Achieve Goals},
  author = {Kaelbling, Leslie Pack},
  booktitle = {Proceedings of the Thirteenth International Joint Conference on Artificial Intelligence},
  year = {1993},
  url = {https://people.csail.mit.edu/lpk/publications.html}
}

@inproceedings{kumar2022should,
  title = {Should I Run Offline Reinforcement Learning or Behavioral Cloning?},
  author = {Kumar, Aviral and Hong, Joey and Singh, Anikait and Levine, Sergey},
  booktitle = {International Conference on Learning Representations},
  year = {2022},
  url = {https://openreview.net/forum?id=AP1MKT37rJ}
}

@inproceedings{singh2021chaining,
  title = {Chaining Behaviors from Data with Model-Free Reinforcement Learning},
  author = {Singh, Avi and Yu, Albert and Yang, Jonathan and Zhang, Jesse and Kumar, Aviral and Levine, Sergey},
  booktitle = {Proceedings of the 2020 Conference on Robot Learning},
  year = {2021},
  volume = {155},
  series = {Proceedings of Machine Learning Research},
  pages = {2162--2177},
  publisher = {PMLR},
  url = {https://proceedings.mlr.press/v155/singh21a.html}
}

@inproceedings{venkatraman2015improving,
  title = {Improving Multi-Step Prediction of Learned Time Series Models},
  author = {Venkatraman, Arun and Hebert, Martial and Bagnell, J. Andrew},
  booktitle = {Proceedings of the AAAI Conference on Artificial Intelligence},
  year = {2015},
  volume = {29},
  pages = {3024--3030},
  doi = {10.1609/aaai.v29i1.9590},
  url = {https://ojs.aaai.org/index.php/AAAI/article/view/9590}
}

@inproceedings{pathak2018zeroshot,
  title = {Zero-Shot Visual Imitation},
  author = {Pathak, Deepak and Mahmoudieh, Parsa and Luo, Guanghao and Agrawal, Pulkit and Chen, Dian and Shentu, Yide and Shelhamer, Evan and Malik, Jitendra and Efros, Alexei A. and Darrell, Trevor},
  booktitle = {International Conference on Learning Representations},
  year = {2018},
  url = {https://pathak22.github.io/zeroshot-imitation/resources/iclr18.pdf}
}

@inproceedings{ho2020denoising,
  title = {Denoising Diffusion Probabilistic Models},
  author = {Ho, Jonathan and Jain, Ajay N. and Abbeel, Pieter},
  booktitle = {Advances in Neural Information Processing Systems},
  year = {2020},
  volume = {33},
  url = {https://proceedings.neurips.cc/paper/2020/hash/4c5bcfec8584af0d967f1ab10179ca4b-Abstract.html}
}

@inproceedings{swamy2026likelihood,
  title = {All Roads Lead to Likelihood: The Value of Reinforcement Learning in Fine-Tuning},
  author = {Swamy, Gokul and Choudhury, Sanjiban and Sun, Wen and Wu, Steven and Bagnell, Drew},
  booktitle = {International Conference on Learning Representations},
  year = {2026},
  url = {https://proceedings.iclr.cc/paper_files/paper/2026/hash/e77d684aae157abd84df1eeb76d8b9cd-Abstract-Conference.html}
}

@inproceedings{wu2025forewarn,
  title = {From Foresight to Forethought: {VLM}-In-the-Loop Policy Steering via Latent Alignment},
  author = {Wu, Yilin and Tian, Thomas and Swamy, Gokul and Bajcsy, Andrea},
  booktitle = {Proceedings of Robotics: Science and Systems},
  year = {2025},
  doi = {10.15607/RSS.2025.XXI.076},
  url = {https://www.roboticsproceedings.org/rss21/p076.html}
}

@inproceedings{ren2024hybrid,
  title = {Hybrid Inverse Reinforcement Learning},
  author = {Ren, Juntao and Swamy, Gokul and Wu, Steven and Bagnell, Drew and Choudhury, Sanjiban},
  booktitle = {Proceedings of the 41st International Conference on Machine Learning},
  year = {2024},
  volume = {235},
  series = {Proceedings of Machine Learning Research},
  pages = {42428--42448},
  publisher = {PMLR},
  url = {https://proceedings.mlr.press/v235/ren24c.html}
}

@article{tong2018neyman,
  title = {Neyman--Pearson classification algorithms and {NP} receiver operating characteristics},
  author = {Tong, Xin and Feng, Yang and Li, Jingyi Jessica},
  journal = {Science Advances},
  year = {2018},
  volume = {4},
  number = {2},
  pages = {eaao1659},
  doi = {10.1126/sciadv.aao1659},
  url = {https://pmc.ncbi.nlm.nih.gov/articles/PMC5804623/}
}

@inproceedings{ke2024ccil,
  title     = {{CCIL}: Continuity-Based Data Augmentation for Corrective Imitation Learning},
  author    = {Ke, Liyiming and Zhang, Yunchu and Deshpande, Abhay and Srinivasa, Siddhartha and Gupta, Abhishek},
  booktitle = {International Conference on Learning Representations},
  year      = {2024},
  url       = {https://openreview.net/forum?id=LQ6LQ8f4y8}
}

@inproceedings{zhang2024dmd,
  title     = {Diffusion Meets {DAgger}: Supercharging Eye-in-hand Imitation Learning},
  author    = {Zhang, Xiaoyu and Chang, Matthew and Kumar, Pranav and Gupta, Saurabh},
  booktitle = {Proceedings of Robotics: Science and Systems},
  address   = {Delft, Netherlands},
  year      = {2024},
  doi       = {10.15607/RSS.2024.XX.048}
}

@article{simeoni2025dinov3,
  title={Dinov3},
  author={Sim{\'e}oni, Oriane and Vo, Huy V and Seitzer, Maximilian and Baldassarre, Federico and Oquab, Maxime and Jose, Cijo and Khalidov, Vasil and Szafraniec, Marc and Yi, Seungeun and Ramamonjisoa, Micha{\"e}l and others},
  journal={arXiv preprint arXiv:2508.10104},
  year={2025}
}
\bibliographystyle{iclr2027_conference}

\clearpage
\appendix
\section{Datasets and Task Setups}
\label{app:datasets}

Below we provide more detail about our real-world tasks, evaluation procedure, and dataset composition in both real and sim. 

\subsection{Real-Robot Tasks}

All demonstrations are collected with handheld UMI devices, specifically iPhUMI~\citep{patel2026behavior}, and policies are given only RGB wrist-camera views with relative proprioception from the starting position of each trajectory. There is no shared-coordinate reference frame, and thus no absolute proprioception that can be used/shared across demonstrations.

\textbf{Sweater folding.}
In this task, the robot first folds each sleeve toward the center of the sweater, then folds the bottom upward. This task tests whether the policy can complete a sequence of coordinated manipulations of a deformable object, and is a longer-horizon task where mistakes may compound. We collect a total of 108 successful trajectories, and 78 failure trajectories for this task.

\textbf{Dish racking.}
In this task, the robot picks up a dish with its left hand, moves towards the center for a handover and reorientation to the right hand, then moves to place the plate into the rack. The handoff requires the receiving grasp to support the dish before the other gripper releases it, and also reorients the plate in the gripper for insertion into the rack to be physically feasible. This task requires high precision: the initial pickup of the thin rim around the plate, the handover coordination, and the insertion between poles not much wider than the plate itself. We collect a total of 179 successful trajectories and 59 failure trajectories for this task. 

\textbf{Evaluation.}
We deploy all policies on a bimanual ARX-5 arm setup. For fairness, we fix 50 starting states per task, which we give to all methods. We use image overlays with the two arms at the same starting pose to reset each object to a precise pose.

\subsection{Simulation Tasks}

\begin{figure*}[t]
    \centering
    \includegraphics[width=\textwidth]{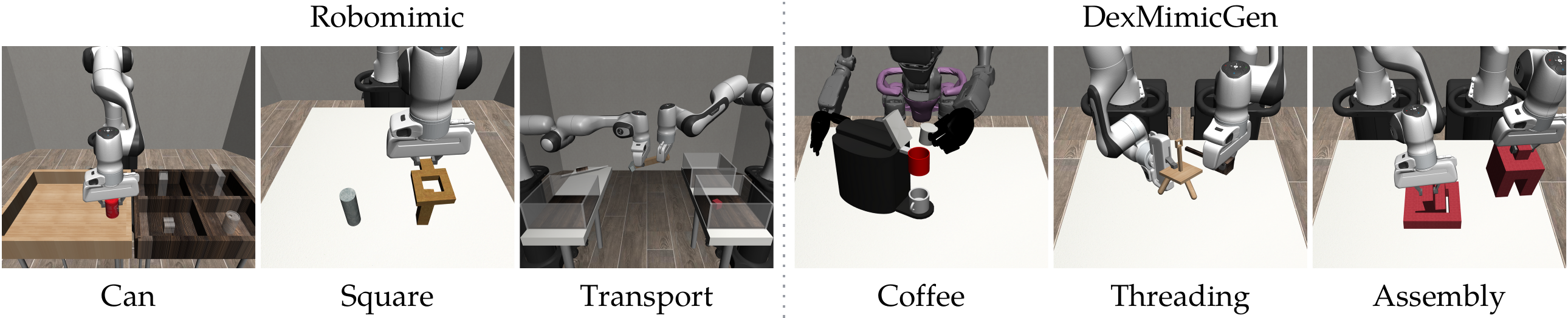}
    \caption{\textbf{Simulation benchmarks.} We evaluate dataset rewriting on three Robomimic tasks with suboptimal demonstrations~\citep{mandlekar2022robomimic} and three DexMimicGen tasks with near-optimal demonstrations~\citep{jiang2025dexmimicgen}.}
    \label{fig:sim_benchmark_tasks}
    \vspace{-4mm}
\end{figure*}

\textbf{Robomimic.}
For each Multi-Human task, we subsample 40 successful demonstrations, with 50\% from the most suboptimal pool and 50\% from the most optimal pool. The minimum/median/maximum trajectory lengths are 98/227/1,050 steps on \textit{Can}, 123/311/1,051 on \textit{Square}, and 392/676.5/2,614 on \textit{Transport}. All methods receive the same selected demonstrations; methods that use failure data also receive the same fixed set of 150 unsuccessful policy rollouts per task. All unsuccessful policy rollouts were generated by three randomly selected Diffusion Policy checkpoints, trained with different configuration than the baselines reported in all experiments. 

\textbf{DexMimicGen.}
We evaluate \textit{Coffee}, \textit{Three Piece Assembly}, and \textit{Threading} using 100 successful demonstrations per task, with the same additional 200 unsuccessful episodes available to methods that use failure data. These demonstrations provide a comparison with the suboptimal Robomimic mixture: the recorded successful behavior is already near-optimal, leaving less opportunity to shorten its continuation.

Following prior evaluation practice~\citep{chi2023diffusionpolicy}, we evaluate checkpoints every 25 epochs on the same 50 fixed evaluation initial conditions and report the mean and standard error of the maximum success rate observed across seeds. All methods use the same checkpoint-evaluation frequency and the same fixed evaluation starts. We tune the baselines extensively rather than relying on their released configurations; Figure~\ref{fig:baseline_comparison} shows that our tuning substantially improves their performance.

\section{Additional Results and Ablations}
\label{app:additional_results}

\subsection{Does Rewriting Reduce Retraced Motion on the Real Robot?}
\label{app:real_robot_retracing}

\begin{figure}[h]
    \centering
    \includegraphics[width=0.48\textwidth]{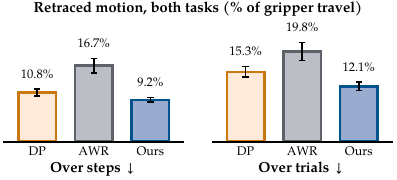}
    \caption{\textbf{Retraced motion on the real robot.} Fraction of gripper travel revisiting a previously exited 2\,cm region, over 100 trials per policy (50 per task). Left: pooled over steps. Right: averaged over trials. Bars show the estimate and standard error.}
    \label{fig:retracing}
\end{figure}

\textbf{\method{} retraces less gripper travel across the two tasks.} We measure the fraction of gripper travel that re-enters a 2\,cm region previously visited and subsequently left during the same rollout, using the same 50 evaluation trials per task and policy as the main results. Pooled over travel across both tasks, this fraction is 9.2\% for \method{}, compared with 10.8\% for DP and 16.7\% for AWR (Figure~\ref{fig:retracing}). Travel pooling gives longer paths more weight; averaging the fraction over trials provides a complementary summary.

\subsection{Can stitching improve with a subset of stages?}
\label{app:robomimic_ablations}
\label{app:stage_ablation}

\begin{figure}[h]
    \centering
    \includegraphics[width=\linewidth]{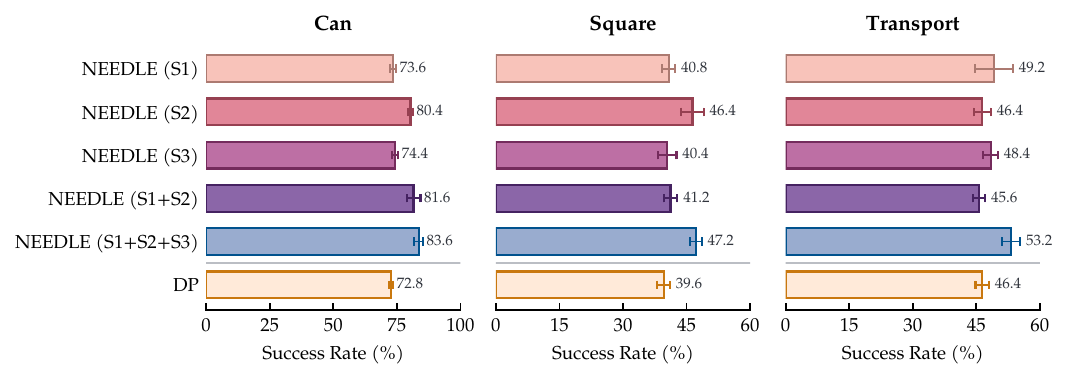}
    \caption{\textbf{Which connections help the policy?} Stage subsets compared with the main-paper DP and \method{} references. Bars show mean success and standard error over five training seeds.}
    \label{fig:stage_ablation}
\end{figure}

The full design draws on three kinds of connections: skipping ahead within a successful demonstration (S1), joining a different successful demonstration (S2), and connecting a failed attempt to a successful demonstration (S3). We seek to better understand the contributions from training on subsets of this combination. Figure~\ref{fig:stage_ablation} shows how well a policy learns with bridges when trained on each stage individually, and from S1+S2 without failure-to-success bridges. We select these subsets from the same bridge pool and retain the original successful demonstrations. 

No single stage reaches the three-stage configuration on all three tasks. Cross-demonstration bridges alone (S2) reach 80.4\% on \textit{Can} and 46.4\% on \textit{Square}. Adding failure-to-success bridges to S1+S2 raises success from 41.2\% to 47.2\% on \textit{Square} and from 45.6\% to 53.2\% on \textit{Transport}. Thus, the relative value of the stages differs across tasks; these results do not imply that each stage is necessary everywhere.

\subsection{What is the best way to train on stitched data?}
\label{app:sampling_ablation}

\begin{figure}[h]
    \centering
    \includegraphics[width=\linewidth]{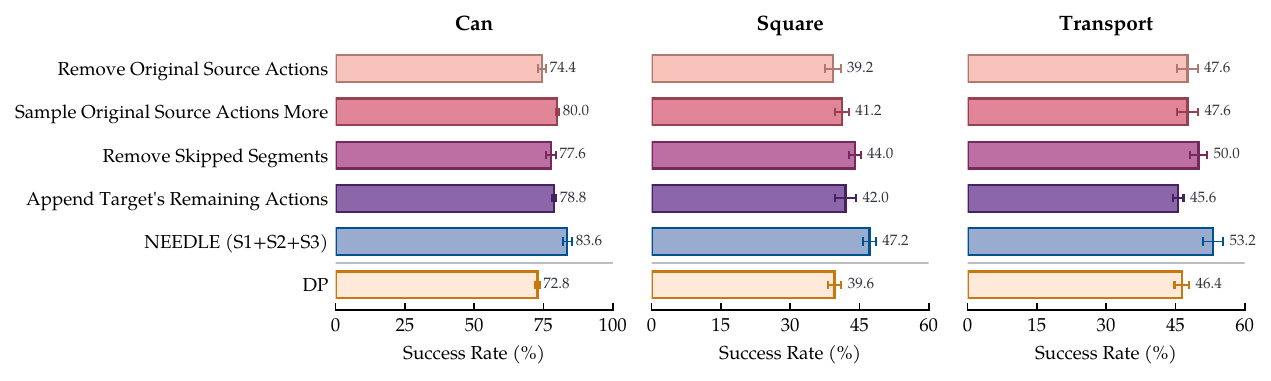}
    \caption{\textbf{How should a policy learn from a bridge?} Rows below \method{} name alternative ways to construct training examples. Bars show mean success and standard error over five training seeds; DP and \method{} are main-paper references.}
    \label{fig:sampling_ablation}
\end{figure}

A verified bridge provides a new action continuation from a recorded observation, but it does not create a complete synthetic trajectory. In particular, its endpoint may differ slightly from the recorded target, or be from a different coordinate system, and taking the shortcut does not eliminate the possibility that the deployed policy will visit states along the original trajectory. Our training construction therefore (i) retains the original actions alongside the bridge, (ii) preserves supervision from bypassed segments, and (iii) stops the new action target at the bridge endpoint rather than concatenating the target trajectory as if the connection were exact. Figure~\ref{fig:sampling_ablation} ablates these choices using the same IDM and verifier as trained in the main paper. We also provide detailed visualizations of the key differences between these sampling methods on our website. 

\textbf{Keep both ways of acting from the bridge's starting point.} Our design retains them as another example of how to act from the same observation. The first alternative removes this example, leaving only the bridge. The second keeps it but doubles its relative sampling weight (0.5 to 1.0), making the recorded actions more common relative to bridge actions. These rows distinguish keeping useful original behavior from simply showing it more often.

\textbf{Keep learning from skipped segments.} A shortcut does not teach the policy what to do at the observations along the skipped segment. We retain those original observation--action examples so the policy can still learn from them. The alternative removes them. It reaches 50.0\% on \textit{Transport}, compared with the 53.2\% reference, with standard errors of 1.9 and 2.2 points respectively.

\textbf{Stop the training target at the bridge endpoint.} We mask action loss when a training window includes both bridge actions and the target demonstration's actions, and learn the target demonstration's remaining actions separately, paired with its own observations. The alternative appends those actions immediately after the bridge, treating the connection as exact. This is possible in simulation's shared coordinate frame, but may be unreliable when a bridge lands slightly away from its target. Even so, the alternative reaches 42.0\% on \textit{Square} and 45.6\% on \textit{Transport}, below their reference values of 47.2\% and 53.2\%.

\subsection{Do the bridge actions themselves matter?}
\label{app:bridge_ablation}

\begin{figure}[h]
    \centering
    \includegraphics[width=\linewidth]{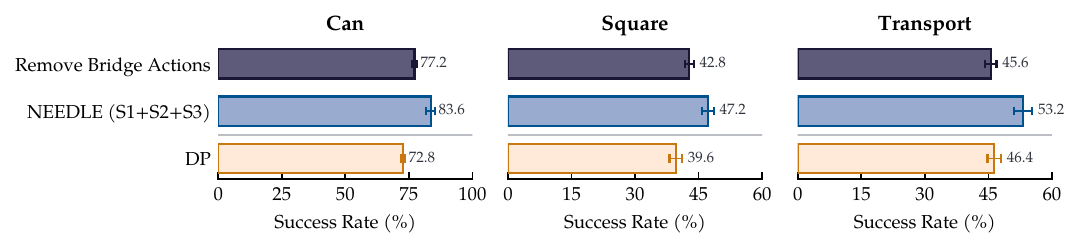}
    \caption{\textbf{Do the bridge actions matter?} The first row removes every training window that contains bridge actions and keeps the reweighted original windows. Bars show mean success and standard error over five training seeds; DP and \method{} are main-paper references.}
    \label{fig:bridge_ablation}
\end{figure}

Removing every training window that contains bridge actions, while keeping the reweighted original windows, lowers success (Figure~\ref{fig:bridge_ablation}) from 83.6\% to 77.2\% on \textit{Can}, from 47.2\% to 42.8\% on \textit{Square}, and from 53.2\% to 45.6\% on \textit{Transport}, so a large part of the gain over DP comes from the bridge actions rather than from reweighting the recorded data.

\subsection{Full DexMimicGen Results}
\label{app:full_dexmimicgen_results}

Table~\ref{tab:dexmimicgen_success_full} expands the near-optimal-demonstration comparison to all evaluated baselines on the three DexMimicGen tasks. All methods draw from a fixed corpus of 100 successful demonstrations; methods that use unsuccessful data receive the same additional 200 failure episodes.

\method{} has the highest mean success on \textit{Coffee} and \textit{Three Piece Assembly}, while AWR performs best on \textit{Threading}. Together with the DP comparison in the main text, these results show that rewriting can still provide modest gains when demonstrations are near-optimal, although the benefit is smaller and task-dependent.

\begin{table*}[h]
    \centering
    \small
    \setlength{\tabcolsep}{6pt}
    \renewcommand{\arraystretch}{1.08}
    \begin{tabular}{lcccccc}
        \toprule
        Task & DP & SBR & AWR & IDQL & SARM & \textbf{Ours} \\
        \midrule
        Coffee & 62.0\,\tabsem{1.2} & 60.7\,\tabsem{2.4} & 52.7\,\tabsem{1.8} & 49.3\,\tabsem{1.8} & 39.3\,\tabsem{1.8} & \textbf{63.3}\,\tabsem{0.7} \\
        Three Piece Assembly & 51.3\,\tabsem{0.7} & 44.7\,\tabsem{1.8} & 47.3\,\tabsem{1.3} & 43.3\,\tabsem{3.5} & 26.7\,\tabsem{1.8} & \textbf{55.3}\,\tabsem{1.8} \\
        Threading & 46.0\,\tabsem{1.2} & 40.7\,\tabsem{2.4} & \textbf{50.0}\,\tabsem{1.2} & 40.0\,\tabsem{1.2} & 34.0\,\tabsem{2.0} & 46.0\,\tabsem{1.2} \\
        \bottomrule
    \end{tabular}
    \caption{\textbf{Full baseline comparison on DexMimicGen.} Success rates (\%) reported as mean and standard error over three training seeds with 50 evaluation initial conditions per seed. \method{} has the highest mean success on two tasks; AWR leads on Threading.}
    \label{tab:dexmimicgen_success_full}
\end{table*}

%
%

\section{Baseline Adaptations and Tuning}
\label{app:baseline_tuning}

\subsection{Shared Setup}
\begin{figure}[h]
    \centering
    \includegraphics[width=\linewidth]{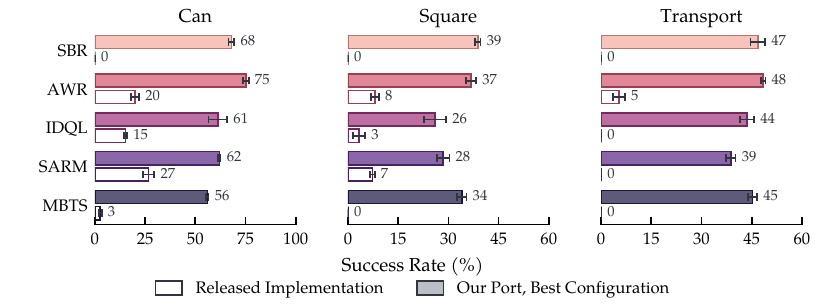}
    \caption{\textbf{Adapting baselines to visual robot control.} Hollow bars show released-policy implementations evaluated on cached DINOv3 features and proprioception; filled bars show the adapted and tuned configurations used in Figure~\ref{fig:robomimic_main_results}. Adaptation improves success across tasks. Error bars show standard errors over three training seeds for original baseline implementations, and five training seeds for our adaptation; zero-height bars indicate measured zero success.}
    \label{fig:baseline_comparison}
\end{figure}

We adapt each baseline to the diffusion-policy backbone used by \method{} and DP, while retaining its data-selection, weighting, or action-reranking mechanism. The adapted policies share cached DINOv3 features~\citep{simeoni2025dinov3}, proprioception, and prediction and execution horizons. Figure~\ref{fig:baseline_comparison} shows that these adaptations substantially improve success over the released-policy implementations evaluated on the same observations. The comparison includes both changes to the policy architecture and task-specific tuning; it assesses the baselines in our visual-control setting.

\subsection{Baseline-Specific Adaptations}

\textbf{SBR.}
SBR retrieves transitions from unsuccessful demonstrations using their similarity to successful demonstrations and weights the retrieved data for policy training. We vary the retrieval representation (raw DINOv3 features or learned autoencoder latents), weighting rule (flat or one-stream), inclusion of proprioception in matching, and use of the autoencoder's internal validation split. All selected configurations remove this internal split, which otherwise excludes four of the 40 successful demonstrations from policy training. Raw-feature retrieval returns no failure transitions on Can and Transport; the selected configurations use autoencoder latents on these tasks.

\textbf{AWR.}
AWR~\citep{peng2019awr} weights policy-training windows by $\min(\exp(\beta(Q-V)), 100)$, using advantages from an IQL critic~\citep{kostrikov2022iql}. We apply this weighting rule to the shared diffusion actor in place of the released Gaussian MLP actor. The critic is trained once and frozen, and the same critic checkpoints are used by IDQL. We compare IQL's antmaze temperature $\beta=10$ (IDQL's default is 3) with temperatures calibrated to target effective sample-size (ESS) fractions of 0.5 or 0.8, computed over per-timestep advantages of either the full dataset or the successful demonstrations. We also vary the actor and critic training corpora independently between successful demonstrations alone and the combined success and failure dataset. The selected actors use successful demonstrations only; the selected critics use all episodes on Can and Transport and successful demonstrations only on Square.

\textbf{IDQL.}
IDQL uses a critic to rerank action samples from a diffusion policy. We vary the number of candidates (1, 16, or 64), selection rule (maximum Q-value or expectile resampling), critic-checkpoint selection rule, and actor and critic training corpora. The selected configuration on all three tasks uses 16 candidates, maximum-Q selection, a success-only actor, and the critic checkpoint with the highest held-out ranking accuracy. We additionally evaluate the same actor without reranking to assess the contribution of critic-based selection.

\textbf{SARM.}
SARM weights policy-training windows using learned visual progress. Its original formulation uses annotated subtask boundaries and language embeddings, which are unavailable in our experimental setting. We therefore evaluate SARM, an adaptation without language conditioning, using either whole-episode progress prediction or a stage-aware model with proportional temporal boundaries. We compare these models with the released rewind augmentation. The selected configurations use the stage-aware model on Can and rewind augmentation on Square and Transport. Proportional boundaries provide temporal structure rather than semantic subtask annotations, so these results characterize the adapted method under our supervision constraints.

\textbf{MBTS.}
MBTS rewrites the offline dataset before policy training: for each transition it retrieves nearby states from other trajectories, proposes a joining action with a learned one-step inverse model, and accepts the splice only if learned dynamics, a density check, and a value-improvement gate all pass, repeating this for several passes. We replace its privileged simulator state (unavailable in our setup) with the same frozen DINOv3 features and proprioception the other baselines use, and reward is the sparse episode-success label rather than a simulator reward. We vary the actor class (the released deterministic MLP versus our diffusion backbone) and train the diffusion actor on the five-pass rewritten corpus. The reported configuration on all three tasks is that rewritten corpus.

\section{Additional Implementation Details}
\label{app:additional_implementation_details}
\label{app:method_details}

\subsection{Shared Architecture and Optimization}
\label{app:shared_training}

\textbf{Visual representation.}
Across Robomimic and real-robot experiments, we use a frozen DINOv3 ViT-B/16 encoder~\citep{simeoni2025dinov3} for the downstream diffusion policy, baselines, IDM, and verifier. The optimizer settings and observation and action horizons in Table~\ref{tab:method_hyperparameters} are also shared across these domains. The shared encoder provides visual features for both learning the recorded behavior and assessing candidate bridges, and allows us to cache image features during training. The IDM and downstream policy pool the image patches into spatial features, while the verifier retains a grid of tokens so that it can attend to different parts of the source and target images.

\begin{table}[ht]
    \centering
    \small
    \begin{tabularx}{\linewidth}{@{}lX@{}}
        \toprule
        Setting & Value \\
        \midrule
        \multicolumn{2}{l}{\textit{Policy, IDM, and verifier}} \\
        Image encoder / resolution & Frozen DINOv3 ViT-B/16 / $224\times224$ \\
        Optimizer / learning rate & AdamW / $10^{-4}$ \\
        Weight decay & $10^{-6}$ \\
        Adam coefficients / epsilon & $(0.95,0.999)$ / $10^{-8}$ \\
        Learning-rate schedule / warmup & Cosine decay / 500 steps \\
        \midrule
        \multicolumn{2}{l}{\textit{Diffusion policy and IDM}} \\
        Observation history / prediction horizon & 2 / 24 steps \\
        Training noise steps & 50 \\
        \midrule
        \multicolumn{2}{l}{\textit{Downstream diffusion policy}} \\
        Action execution horizon & 12 steps \\
        Batch size & 64 \\
        Training epochs (Robomimic / real robot) & 300 / 500 \\
        \bottomrule
    \end{tabularx}
    \caption{\textbf{Shared visual and training settings.} The encoder and optimizer settings apply to the policy, IDM, and verifier; the remaining rows specify their scope.}
    \label{tab:method_hyperparameters}
\end{table}

\subsection{Inverse Dynamics Model}
\label{app:idm_details}

\textbf{Prediction and execution horizons.}
The 24-row action prediction and the 23-step bridge horizon refer to different parts of the same sequence. With two observation steps, one predicted row aligns with the earlier observation, leaving 23 actions from the current source onward. The downstream policy executes 12 actions before replanning. For IDM training, the remaining rows after reaching a sampled goal hold the target pose; they do not continue along the demonstration past that goal. Table~\ref{tab:idm_hyperparameters} collects the architecture and training settings.

\begin{table}[h]
    \centering
    \small
    \setlength{\tabcolsep}{5pt}
    \begin{tabularx}{\linewidth}{@{}lX@{}}
        \toprule
        Setting & Value \\
        \midrule
        \multicolumn{2}{l}{\textit{Architecture}} \\
        Visual feature pooling & Spatial softmax over a $14\times14$ patch grid \\
        DDIM sampling steps (model configuration) & 50 \\
        Maximum executable bridge length & 23 actions \\
        Action representation & Absolute pose (simulation); source-relative pose (real robot); 6D rotation and gripper command \\
        U-Net channel widths & 512, 1024, 2048 \\
        Diffusion-step embedding dimension & 256 \\
        U-Net kernel size / normalization groups & 5 / 8 \\
        Goal-conditioning MLP hidden widths & 256, 256 \\
        \midrule
        \multicolumn{2}{l}{\textit{Optimization}} \\
        Batch size & 128 \\
        \bottomrule
    \end{tabularx}
    \caption{\textbf{IDM architecture and hyperparameters.} Shared visual and optimizer settings are in Table~\ref{tab:method_hyperparameters}.}
    \label{tab:idm_hyperparameters}
\end{table}

\subsection{Verifier}
\label{app:verifier_details}

The verifier learns from positive source--goal pairs connected by recorded actions and negative pairs that those actions do not connect. Negative examples include goals beyond the action horizon and in other trajectories. Sampling determines which examples the verifier sees, while the loss assigns equal total weight to the positive and negative terms. Increasing the number of negatives per positive therefore does not increase their aggregate loss weight. Table~\ref{tab:verifier_sampling} collects the architecture, sampling, loss, and calibration settings.

\begin{table}[ht]
    \centering
    \small
    \setlength{\tabcolsep}{5pt}
    \begin{tabularx}{\linewidth}{@{}lX@{}}
        \toprule
        Setting & Value \\
        \midrule
        \multicolumn{2}{l}{\textit{Architecture}} \\
        Source and target visual tokens & $7\times7$ patch grid per camera \\
        Transformer layers / attention heads / hidden width & 2 / 4 / 512 \\
        Feed-forward width / dropout & 4096 / 0 \\
        Attention / positional encoding & Causal / rotary \\
        Proprioception / action normalization & Source only / per action dimension \\
        Maximum scored action prefix & 23 actions \\
        \midrule
        \multicolumn{2}{l}{\textit{Training pairs and loss}} \\
        Positive anchors / negatives per group & 1 / 7 \\
        Spacing between candidate targets in one episode & 23 steps \\
        Loss function & Balanced binary cross-entropy \\
        Positive / negative aggregate loss weight & 0.5 / 0.5 \\
        \midrule
        \multicolumn{2}{l}{\textit{Optimization and calibration}} \\
        Batch size / training epochs & 1024 / 40 \\
        Initial calibration target false-positive rate & 0.01 \\
        \bottomrule
    \end{tabularx}
    \caption{\textbf{Verifier architecture and hyperparameters.} Sampling and training values describe the Robomimic runs. Shared visual and optimizer settings are in Table~\ref{tab:method_hyperparameters}.}
    \label{tab:verifier_sampling}
\end{table}

\subsection{Stitching and Training-Window Construction}
\label{app:stitching_details}

\textbf{Why keep one bridge per source?}
A source state can have several candidate targets. Keeping all of them would make that state's contribution depend on how many matches happened to be found and could give the policy several competing action targets at the same departure. We instead retain the shortest bridge-plus-suffix path for each source. For successful sources, this also maximizes the planned reduction in remaining steps, which must be positive. This separates choosing a useful continuation from deciding how often to train on it.

\begin{table}[h]
    \centering
    \small
    \setlength{\tabcolsep}{5pt}
    \begin{tabularx}{\linewidth}{@{}p{0.60\linewidth}X@{}}
        \toprule
        Setting & Value \\
        \midrule
        \multicolumn{2}{l}{\textit{Candidate search and acceptance}} \\
        IDM proposals per pair & 8 \\
        Bridges retained per source & 1 \\
        Successful-source preference & Shortest bridge-plus-suffix path; positive steps saved \\
        Failure-source preference & Shortest bridge-plus-suffix path \\
        \midrule
        \multicolumn{2}{l}{\textit{Relative sampling weights per window}} \\
        Other recorded windows & 0.5 \\
        Recorded windows skipped by a within-trajectory bridge & 0.5 \\
        Recorded action window at a bridge source & 0.5 \\
        Recorded suffix after the target (S1 / S2) & 1.0 / 1.0 \\
        Recorded failure-prefix windows & 0 \\
        \bottomrule
    \end{tabularx}
    \caption{\textbf{Stitch generation and training-window construction.} Shared settings for the nominal Robomimic runs; task-specific candidate budgets appear in Table~\ref{tab:robomimic_hyperparameters}. Weights control how often a window is sampled, rather than scaling its action loss.}
    \label{tab:stitching_hyperparameters}
\end{table}

Table~\ref{tab:robomimic_hyperparameters} lists the task-specific candidate budgets used for the main Robomimic comparison.

\begin{table}[h]
    \centering
    \small
    \setlength{\tabcolsep}{4pt}
    \begin{tabularx}{\linewidth}{@{}lXXX@{}}
        \toprule
        Setting & Can & Square & Transport \\
        \midrule
        \multicolumn{4}{l}{\textit{Observations, data, and evaluation}} \\
        RGB cameras & External + wrist & External + wrist & 2 shoulder + 2 wrist \\
        Action dimensions & 10 & 10 & 20 \\
        Successful / unsuccessful episodes & {40 / 150} & {40 / 150} & {40 / 150}  \\
        Full evaluation budget & 500 & 500 & 700 \\
        Evaluation starts per seed & 50 & 50 & 50 \\
        \midrule
        \multicolumn{4}{l}{\textit{Task-specific stitching choices}} \\
        IDM training epochs & 300 & 150 & 150 \\
        Candidate cap, each of S1 and S2 & $10^5$ & $10^5$ & $10^5$ \\
        Failure-candidate cap & $10^6$ & $10^5$ & $10^5$ \\
        Failure targets per source & 16 & 8 & 8 \\
        \bottomrule
    \end{tabularx}
    \caption{\textbf{Robomimic task settings.} Columns describe the nominal policies used for the main success comparison. Shared training settings appear in Table~\ref{tab:method_hyperparameters}.}
    \label{tab:robomimic_hyperparameters}
\end{table}

\subsection{Compute Resources}
\label{app:compute_resources}

All runs use a single NVIDIA RTX PRO 6000 Blackwell Server Edition GPU unless noted; earlier development runs also used NVIDIA L40S GPUs. In simulation, the IDM trains for 300 epochs on \textit{Can} and 150 epochs on \textit{Square} and \textit{Transport} in 2--4 h, the verifier for 40 epochs in 6--11 GPU-hours, offline stitching of the full candidate pool takes about 2 h, and the downstream policy trains for 300 epochs with periodic simulator evaluation in 2--9 h depending on the task. The IDM and verifier can be trained in parallel on separate GPUs. Baseline policies cost the same as our downstream policy since they share the training recipe.

\subsection{Evaluation and Diagnostic Protocols}
\label{app:evaluation_details}

\textbf{Policy success and completion budgets.}
Robomimic policies are trained for 300 epochs and evaluated every 25 epochs on 50 initial conditions, paired across baselines. Figure~\ref{fig:robomimic_main_results} reports success and standard error over five training seeds evaluated for 50 episodes each; completion steps are measured at that same evaluation point and averaged over successful rollouts. For the rollout panels in Figure~\ref{fig:stitching_completion}, we evaluate the selected policies from 50 matched held-out initial conditions per task and training seed. A trial contributes to success at budget $b$ only if its first success occurs by step $b$. Full evaluation budgets are 500 steps for \textit{Can} and \textit{Square} and 700 for \textit{Transport}, which we take directly from the Multi-Human setting in the Robomimic benchmark~\citep{mandlekar2022robomimic}.

\textbf{Bridge continuations.}
For the continuation panels in Figure~\ref{fig:stitching_completion}, we restore the same simulator state for each source state that ends up with a proposed stitch, execute either the recorded actions or the selected bridge for the bridge's duration, and resume the same DP policy from each landing. We measure remaining steps from the end of this action sequence. The comparison uses the fastest successful resume; the fitted distributions therefore describe completion from jointly successful resumes.

\textbf{Recovery from failure states.}
The rollout panels in Figure~\ref{fig:stitching_recovery} use 50 sampled failure states per task, shared across methods and evaluated with five training seeds and 16 resumes per state. We first compute the success fraction across the 16 resumes from each state, then average over states and training seeds. We focus on failure states in which the object remains in the robot's reachable workspace, excluding cases such as Transport objects dropped to the ground. Each resume receives a full task horizon.

\end{document}